\documentclass{article}

\PassOptionsToPackage{numbers, compress}{natbib}
\usepackage[final]{neurips_2020}

\usepackage[utf8]{inputenc} 
\usepackage[T1]{fontenc}    
\usepackage[colorlinks]{hyperref}       
\usepackage{url}            
\usepackage{booktabs}       
\usepackage{amsfonts}       
\usepackage{nicefrac}       
\usepackage{microtype}      
\usepackage{mathtools,bbm}
\usepackage{color}
\usepackage{multirow}
\usepackage{caption}
\usepackage{subcaption}
\usepackage{makecell}
\usepackage{wrapfig}
\usepackage{cleveref}
\usepackage{textcomp}
\usepackage{gensymb}
\usepackage{marginnote}
\crefname{equation}{}{}
\definecolor{BrickRed}{rgb}{0.6,0,0}
\definecolor{RoyalBlue}{rgb}{0,0,0.8}
\definecolor{Tdgreen}{rgb}{0,0.4,0.7}
\hypersetup{citecolor=BrickRed, linkcolor=RoyalBlue}

\DeclarePairedDelimiter\abs{\lvert}{\rvert}
\DeclarePairedDelimiter\norm{\lVert}{\rVert}
\newcommand{\R}{\mathbb{R}}
\newcommand{\E}{\mathbb{E}}
\newcommand{\stdv}[1]{\scriptsize$\pm$#1}

\newcommand{\ie}{\textit{i}.\textit{e}.}
\newcommand{\eg}{\textit{e}.\textit{g}.}

\newcommand{\code}[1]{\texttt{#1}}

\newcommand{\reviewer}[3]{
	\expandafter\newcommand\csname #1\endcsname[1]{
		\textcolor{#3}{[#2: ##1]}
	}
}

\title{CSI: Novelty Detection via Contrastive Learning\\on Distributionally Shifted Instances}

\author{%
  Jihoon Tack\thanks{Equal contribution}$\:\:^\dagger$, Sangwoo Mo$^{*\ddagger}$, Jongheon Jeong$^\ddagger$, Jinwoo Shin$^{\dagger\ddagger}$ \\
  $^\dagger$Graduate School of AI, KAIST\\
  $^\ddagger$School of Electrical Engineering, KAIST\\
  \texttt{\{jihoontack,swmo,jongheonj,jinwoos\}@kaist.ac.kr} \\
}

\begin{document}

\maketitle

\begin{abstract}
Novelty detection, \ie, identifying whether a given sample is drawn from outside the training distribution, is essential for reliable machine learning. To this end, there have been many attempts at learning a representation well-suited for novelty detection and designing a score based on such representation. In this paper, we propose a simple, yet effective method named \emph{contrasting shifted instances} (CSI), inspired by the recent success on contrastive learning of visual representations. Specifically, in addition to contrasting a given sample with other instances as in conventional contrastive learning methods, our training scheme contrasts the sample with distributionally-shifted augmentations of itself. Based on this, we propose a new detection score that is specific to the proposed training scheme. Our experiments demonstrate the superiority of our method under various novelty detection scenarios, including unlabeled one-class, unlabeled multi-class and labeled multi-class settings, with various image benchmark datasets. Code and pre-trained models are available at \url{https://github.com/alinlab/CSI}.
\end{abstract}

\section{Introduction}
\label{sec:intro}

Out-of-distribution (OOD) detection \citep{hodge2004survey}, also referred to as a novelty- or anomaly detection is a task of identifying whether a test input is drawn far from the training distribution (in-distribution) or not. In general, the OOD detection problem aims to detect OOD samples where a detector is allowed to access only to training data. The space of OOD samples is typically huge, \ie, an OOD sample can vary significantly and arbitrarily from the given training distribution. Hence, assuming specific prior knowledge, \eg, external data representing some specific OODs, may introduce a bias to the detector. The OOD detection is a classic yet essential problem in machine learning, with a broad range of applications, including medical diagnosis \citep{caruana2015intelligible}, fraud detection \citep{phua2010comprehensive}, and autonomous driving \citep{eykholt2018robust}.

A long line of literature has thus been proposed, including density-based \citep{zhai2016deep,nalisnick2019deep,choi2018waic,nalisnick2019detecting,du2019implicit,ren2019likelihood,serra2020input,grathwohl2020your}, reconstruction-based \citep{schlegl2017unsupervised,zong2018deep,deecke2018anomaly,pidhorskyi2018generative,perera2019ocgan,choi2020novelty}, one-class classifier \citep{scholkopf2000support,ruff2018deep}, and self-supervised \citep{golan2018deep,hendrycks2019using_self,bergman2020classification} approaches. Overall, a majority of recent literature is concerned with (a) modeling the representation to better encode normality \citep{hendrycks2019using_pre,hendrycks2019using_self}, and (b) defining a new detection score \citep{ruff2018deep,bergman2020classification}. In particular, recent studies have shown that inductive biases from self-supervised learning significantly help to learn discriminative features for OOD detection \citep{golan2018deep,hendrycks2019using_self,bergman2020classification}.

Meanwhile, recent progress on self-supervised learning has proven the effectiveness of \emph{contrastive learning} in various domains, \eg, computer vision \citep{he2019momentum,chen2020simple}, audio processing \citep{oord2018representation}, and reinforcement learning \citep{srinivas2020curl}. Contrastive learning extracts a strong inductive bias from multiple (similar) views of a sample by let them attract each other, yet repelling them to other samples.  \emph{Instance discrimination} \citep{wu2018unsupervised} is a special type of contrastive learning where the views are restricted up to different augmentations, which have achieved state-of-the-art results on visual representation learning \citep{he2019momentum,chen2020simple}.

Inspired by the recent success of instance discrimination, we aim to utilize its power of representation learning for OOD detection. To this end, we investigate the following questions: (a) how to learn a (more) discriminative representation for detecting OODs and (b) how to design a score function utilizing the representation from (a). We remark that the desired representation for OOD detection may differ from that for standard representation learning \citep{hendrycks2019using_pre,hendrycks2019using_self}, as the former aims to discriminate in-distribution and OOD samples, while the latter aims to discriminate \textit{within} in-distribution samples. 

We first found that existing contrastive learning scheme 
is already reasonably effective for detecting OOD samples with a proper detection score. We further observe that one can improve its performance by utilizing ``hard'' augmentations, \eg, rotation, that were known to be harmful and unused for the standard contrastive learning \citep{chen2020simple}. In particular, while the
existing contrastive learning schemes act by pulling all augmented samples toward the original sample, we suggest to additionally push the samples with hard or distribution-shifting augmentations away from the original. We observe that contrasting shifted samples help OOD detection, as the model now learns a new task of discriminating \textit{between} in- and out-of-distribution, in addition to the original task of discriminating \textit{within} in-distribution.

\textbf{Contribution.}
We propose a simple yet effective method for OOD detection, coined \emph{contrasting shifted instances} (CSI). Built upon the existing contrastive learning scheme \citep{chen2020simple}, we propose two novel additional components: (a) a new training method which contrasts distributionally-shifted augmentations (of the given sample) in addition to other instances, and (b) a score function which utilizes both the contrastively learned representation and our new training scheme in (a). Finally, we show that CSI enjoys broader usage by applying it to improve the confidence-calibration of the classifiers: it relaxes the overconfidence issue in their predictions for both in- and out-of-distribution samples while maintaining the classification accuracy.

We verify the effectiveness of CSI under various environments of detecting OOD, including unlabeled one-class, unlabeled multi-class, and labeled multi-class settings. To our best knowledge, we are the first to demonstrate all three settings under a single framework. Overall, CSI outperforms the baseline methods for all tested datasets. In particular, CSI achieves new state-of-the-art results\footnote{We do not compare with methods using \textit{external} OOD samples \citep{hendrycks2019deep,ruff2020deep}.} on one-class classification, \eg, it improves the mean area under the receiver operating characteristics (AUROC) from 90.1\% to 94.3\% (+4.2\%) for CIFAR-10 \citep{krizhevsky2009learning}, 79.8\% to 89.6\% (+9.8\%) for CIFAR-100 \citep{krizhevsky2009learning}, and 85.7\% to 91.6\% (+5.9\%) for ImageNet-30 \citep{hendrycks2019using_self} one-class datasets, respectively. We remark that CSI gives a larger improvement in harder (or near-distribution) OOD samples. To verify this, we also release new benchmark datasets: fixed version of the resized LSUN and ImageNet \citep{liang2018enhancing}.

We remark that learning representation to discriminate in- vs. out-of-distributions is an important but under-explored problem. We believe that our work would guide new interesting directions in the future, for both representation learning and OOD detection.

\vspace{-0.05in}
\section{CSI: Contrasting shifted instances}
\vspace{-0.05in}
\label{sec:method}

For a given dataset $\{x_m\}_{m=1}^M$ sampled from a data distribution $p_{\text{data}}(x)$ on the data space $\mathcal{X}$, the goal of out-of-distribution (OOD) detection is to model a detector from $\{x_m\}$ that identifies whether $x$ is sampled from the data generating distribution (or in-distribution) $p_{\text{data}}(x)$ or not. As modeling $p_{\text{data}}(x)$ directly is prohibitive in most cases, many existing methods for OOD detection define a \emph{score function} $s(x)$ that a high value heuristically represents that $x$ is from in-distribution.

\vspace{-0.05in}
\subsection{Contrastive learning}
\vspace{-0.05in}
\label{sec:method-prelim}

The idea of \emph{contrastive learning} is to learn an encoder $f_\theta$ to extract the necessary information to distinguish similar samples from the others. Let $x$ be a query, $\{x_+\}$, and $\{x_-\}$ be a set of positive and negative samples, respectively, and $\mathrm{sim}(z,z') := z \cdot z' / \norm{z} \norm{z'}$ be the cosine similarity. Then, the primitive form of the \emph{contrastive loss} is defined as follows:
\begin{equation}
    \mathcal{L}_{\texttt{con}} (x, \{x_+\}, \{x_-\}) 
    := - \frac{1}{\abs{\{x_+\}}} 
        \log \frac{\sum_{x' \in \{x_+\}}\exp(\mathrm{sim}(z(x), z(x')) / \tau)}
        {\sum_{x' \in \{x_+\} \cup \{x_-\}} \exp(\mathrm{sim}(z(x), z(x')) / \tau)},
    \label{eq:loss-con}
\end{equation}
where $\abs{\{x_+\}}$ denotes the cardinality of the set $\{x_+\}$, $z(x)$ denotes the output feature of the contrastive layer, and $\tau$ denotes a temperature hyper-parameter. One can define the contrastive feature $z(x)$ directly from the encoder $f_\theta$, \ie, $z(x) = f_\theta(x)$ \citep{he2019momentum}, or apply an additional projection layer $g_\phi$, \ie, $z(x) = g_\phi(f_\theta(x))$ \citep{chen2020simple}. We use the projection layer following the recent studies \citep{chen2020simple,khosla2020supervised}.

In this paper, we specifically consider the simple contrastive learning (\textit{SimCLR}) \citep{chen2020simple}, a simple and effective objective based on the task of \emph{instance discrimination} \citep{wu2018unsupervised}. Let $\tilde{x}^{(1)}_i$ and $\tilde{x}^{(2)}_i$ be two independent augmentations of $x_i$ from a pre-defined family $\mathcal{T}$, namely, $\tilde{x}^{(1)}:=T_1(x_i)$ and $\tilde{x}^{(2)}:=T_2(x_i)$, where $T_1, T_2 \sim \mathcal{T}$. Then the SimCLR objective can be defined by the contrastive loss \eqref{eq:loss-con} where each $(\tilde{x}^{(1)}_i, \tilde{x}^{(2)}_i)$ and $(\tilde{x}^{(2)}_i, \tilde{x}^{(1)}_i)$ are considered as query-key pairs while others being negatives. Namely, for a given batch $\mathcal{B}:=\{x_i\}_{i=1}^B$, the SimCLR objective is defined as follows:
\begin{equation}
    \mathcal{L}_{\texttt{SimCLR}}(\mathcal{B};\mathcal{T})
        := \frac{1}{2B} \sum_{i=1}^B
        \mathcal{L}_{\texttt{con}} (\tilde{x}^{(1)}_i, \tilde{x}^{(2)}_i, \tilde{\mathcal{B}}_{-i})
        + \mathcal{L}_{\texttt{con}} (\tilde{x}^{(2)}_i, \tilde{x}^{(1)}_i, \tilde{\mathcal{B}}_{-i}),
    \label{eq:loss-simclr}
\end{equation}
where $\tilde{\mathcal{B}}:=\{\tilde{x}^{(1)}_i\}_{i=1}^B\cup\{\tilde{x}^{(2)}_i\}_{i=1}^B$ and $\tilde{\mathcal{B}}_{-i}:=\{\tilde{x}^{(1)}_j\}_{j\neq i}\cup\{\tilde{x}^{(2)}_j\}_{j\neq i}$.

\vspace{-0.05in}
\subsection{Contrastive learning for distribution-shifting transformations}
\vspace{-0.05in}
\label{sec:method-train}

\citet{chen2020simple} has performed an extensive study on which family of augmentations $\mathcal{T}$ leads to a better representation when used in SimCLR, \ie, which transformations should $f_\theta$ consider as positives. Overall, the authors report that some of the examined augmentations (\eg, rotation), sometimes degrades the discriminative performance of SimCLR. One of our key findings is that such augmentations can be useful for OOD detection by considering them as \emph{negatives} - contrast from the original sample. In this paper, we explore which family of augmentations $\mathcal{S}$, which we call \emph{distribution-shifting transformations}, or simply \emph{shifting transformations}, would lead to better representation in terms of OOD detection when used as negatives in SimCLR.

\textbf{Contrasting shifted instances.}
We consider a set $\mathcal{S}$ consisting of $K$ different (random or deterministic) transformations, including the identity $I$: namely, we denote $\mathcal{S}:=\{S_0 = I, S_1, \dots, S_{K-1} \}$. In contrast to the vanilla SimCLR that considers augmented samples as positive to each other, we attempt to consider them as negative if the augmentation is from $\mathcal{S}$. For a given batch of samples $\mathcal{B}=\{x_i\}_{i=1}^B$, this can be done simply by augmenting $\mathcal{B}$ via $\mathcal{S}$ before putting it into the SimCLR loss defined in \eqref{eq:loss-simclr}: namely, we define \emph{contrasting shifted instances} (con-SI) loss as follows:
\begin{equation}
    \mathcal{L}_{\texttt{con-SI}}
        := \mathcal{L}_{\texttt{SimCLR}} \left( \bigcup_{S \in \mathcal{S}} \mathcal{B}_S; \mathcal{T} \right),
        \quad \text{where} ~~ \mathcal{B}_S := \{S(x_i)\}_{i=1}^B.
    \label{eq:loss-con-SI}
\end{equation}
Here, our intuition is to regard each distributionally-shifted sample (\ie, $S \neq I$) as an OOD with respect to the original. In this respect, con-SI attempts to discriminate an in-distribution (\ie, $S = I$) sample from other OOD (\ie, $S \in \{S_1,\dots,S_{K-1}\}$) samples. We further verify the effectiveness of con-SI in our experimental results: although con-SI does not improve representation for standard classification, it does improve OOD detection significantly (see linear evaluation in Section \ref{sec:exp-ablation}).

\textbf{Classifying shifted instances.}
In addition to contrasting shifted instances, we consider an auxiliary task that predicts which shifting transformation $y^{S}\in\mathcal{S}$ is applied for a given input $x$, in order to facilitate $f_\theta$ to discriminate each shifted instance.  Specifically, we add a linear layer to $f_\theta$ for modeling an auxiliary softmax classifier $p_\texttt{cls-SI}(y^\mathcal{S}|x)$, as in \citep{golan2018deep,hendrycks2019using_self,bergman2020classification}. Let $\tilde{\mathcal{B}}_S$ be the batch augmented from $\mathcal{B}_S$ via SimCLR; then, we define \textit{classifying shifted instances} (cls-SI) loss as follows:
\begin{equation}
    \mathcal{L}_\texttt{cls-SI}
        := \frac{1}{2B}\frac{1}{K} \sum_{S \in \mathcal{S}} ~\sum_{\tilde{x}_{S} \in \tilde{\mathcal{B}}_S}
        - \log p_\texttt{cls-SI}(y^\mathcal{S} = S \mid \tilde{x}_{S}).
    \label{eq:loss-cls-SI}
\end{equation}

The final loss of our proposed method, \emph{CSI}, is defined by combining the two objectives:
\begin{equation}
    \mathcal{L}_\texttt{CSI} = \mathcal{L}_\texttt{con-SI} + \lambda \cdot \mathcal{L}_\texttt{cls-SI}
    \label{eq:loss-csi}
\end{equation}
where $\lambda > 0$ is a balancing hyper-parameter. We simply set $\lambda = 1$ for all our experiments.

\textbf{\textit{OOD-ness}: How to choose the shifting transformation?}
In principle, we choose the shifting transformation that generates the most OOD-like yet semantically meaningful samples. Intuitively, such samples can be most effective (‘nearby’ but ‘not-too-nearby’) OOD samples, as also discussed in Section~\ref{sec:exp-ablation}. More specifically, we measure the \textit{OOD-ness} of a transformation by the area under the receiver operating characteristics (AUROC) between in-distribution vs. transformed samples under vanilla SimCLR, using the detection score \eqref{eq:score-con} defined in Section~\ref{sec:method-detect}. The transformation with high OOD-ness values (\ie, OOD-like) indeed performs better (see Table~\ref{tab:ood_aug} and Table~\ref{tab:ablation-aug} in Section~\ref{sec:exp-ablation}).

\vspace{-0.05in}
\subsection{Score functions for detecting out-of-distribution}
\vspace{-0.05in}
\label{sec:method-detect}

Upon the representation $z(\cdot)$ learned by our proposed training objective, we define several score functions for detecting out-of-distribution; whether a given $x$ is OOD or not. We first propose a detection score that is applicable to any contrastive representation. We then introduce how one could incorporate additional information learned by contrasting (and classifying) shifted instances as in \eqref{eq:loss-csi}.

\textbf{Detection score for contrastive representation.}
Overall, we find that two features from SimCLR representations are surprisingly effective for detecting OOD samples: (a) the \emph{cosine similarity} to the nearest training sample in $\{x_m\}$, \ie, $\max_m \mathrm{sim}(z(x_m), z(x))$, and (b) the \emph{norm} of the representation, \ie, $\norm{z(x)}$. Intuitively, the contrastive loss increases the norm of in-distribution samples, as it is an easy way to minimize the cosine similarity of identical samples by increasing the denominator of \eqref{eq:loss-con}. We discuss further detailed analysis of both features in Appendix \ref{appx:norm}. We simply combine these features to define a detection score $s_{\texttt{con}}$ for contrastive representation:
\begin{equation}
    s_\texttt{con}(x;\{x_m\}) := \max_m ~ \mathrm{sim}(z(x_m), z(x)) \cdot \norm{z(x)}.
    \label{eq:score-con}
\end{equation}
We also discuss how one can reduce the computation and memory cost by choosing a proper subset (\ie, coreset) of training samples in Appendix \ref{appx:coreset}.

\textbf{Utilizing shifting transformations.}
Given that our proposed $\mathcal{L}_{\texttt{CSI}}$ is used for training, one can further improve the detection score $s_\texttt{con}$ significantly by incorporating shifting transformations $\mathcal{S}$. Here, we propose two additional scores, $s_\texttt{con-SI}$ and $s_\texttt{cls-SI}$, where are corresponded to $\mathcal{L}_{\texttt{con-SI}}$ \eqref{eq:loss-con-SI} and $\mathcal{L}_{\texttt{cls-SI}}$ \eqref{eq:loss-cls-SI}, respectively.

Firstly, we define $s_\texttt{con-SI}$ by taking an expectation of $s_\texttt{con}$ over $S\in\mathcal{S}$:
\begin{equation}
    s_\texttt{con-SI}(x;\{x_m\}) := \sum_{S \in \mathcal{S}} \lambda^\texttt{con}_S ~ s_\texttt{con}(S(x);\{S(x_m)\}),
    \label{eq:score-con-SI}
\end{equation}
where $\lambda^\texttt{con}_S := M / \sum_m s_\texttt{con}(S(x_m); \{S(x_m)\}) = M / \sum_m \norm{z(S(x_m))}$ for $M$ training samples is a balancing term to scale the scores of each shifting transformation (See Appendix \ref{appx:balance} for details).

Secondly, we define $s_\texttt{cls-SI}$ utilizing the auxiliary classifier $p(y^\mathcal{S}|x)$ upon $f_\theta$ as follows:
\begin{align}
    s_\texttt{cls-SI}(x) := \sum_{S \in \mathcal{S}} \lambda^\texttt{cls}_S ~ W_S f_\theta(S(x)),
    \label{eq:score-cls-SI}
\end{align}
where $\lambda^\texttt{cls}_S := M / \sum_m [W_S f_\theta(S(x_m))]$ are again balancing terms similarly to above, and $W_S$ is the weight vector in the linear layer of $p(y^\mathcal{S}|x)$ per $S\in\mathcal{S}$.

Finally, the combined score for CSI representation is defined as follows:
\begin{align}
    s_\texttt{CSI}(x;\{x_m\}) := s_\texttt{con-SI}(x;\{x_m\}) + s_\texttt{cls-SI}(x).
    \label{eq:score-csi}
\end{align}

\textbf{Ensembling over random augmentations.}
In addition, we find one can further improve each of the proposed scores by ensembling it over random augmentations $T(x)$ where $T \sim \mathcal{T}$. Namely, for instance, the \textit{ensembled} CSI score is defined by $s_\texttt{CSI-ens}(x) := \E_{T \sim \mathcal{T}} [s_\texttt{CSI}(T(x))]$. Unless otherwise noted, we use these ensembled versions of  \cref{eq:score-con,eq:score-con-SI,eq:score-cls-SI,eq:score-csi} in our experiments. See Appendix \ref{appx:test_aug} for details.

\vspace{-0.05in}
\subsection{Extension for training confidence-calibrated classifiers}
\vspace{-0.05in}
\label{sec:method-label}

Furthermore, we propose an extension of CSI for training \emph{confidence-calibrated} classifiers \citep{hendrycks2017baseline,lee2018training} from a given labeled dataset $\{(x_m,y_m)\}_m \subseteq \mathcal{X} \times \mathcal{Y}$ by adapting it to \emph{supervised contrastive learning} (SupCLR) \citep{khosla2020supervised}. Here, the goal is to model a classifier $p(y|x)$ that is (a) accurate on predicting $y$ when $x$ is in-distribution, and (b) the \emph{confidence} $s_\texttt{sup}(x) := \max_y p(y|x)$ \citep{hendrycks2017baseline} of the classifier is \emph{well-calibrated}, \ie, $s_\texttt{sup}(x)$ should be low if $x$ is an OOD sample or $\arg\max_y p(y|x)\neq \mbox{true label}$.

\textbf{Supervised contrastive learning (SupCLR).} SupCLR is a supervised extension of SimCLR that contrasts samples in \emph{class-wise}, instead of in instance-wise: every samples of the same classes are considered as positives. Let $\mathcal{C}=\{(x_i,y_i)\}_{i=1}^{B}$ be a training batch with class labels $y_i \in \mathcal{Y}$, and $\tilde{\mathcal{C}}$ be an augmented batch by random transformation $\mathcal{T}$, \ie, $\tilde{\mathcal{C}} := \{(\tilde{x}_j,y_j) \mid \tilde{x}_j \in \tilde{\mathcal{B}}\}$. For a given label $y$, we divide $\tilde{\mathcal{C}}$ into two subsets $\tilde{\mathcal{C}} = \tilde{\mathcal{C}}_{y} \cup \tilde{\mathcal{C}}_{-y}$ where $\tilde{\mathcal{C}}_{y}$ contains the samples of label $y$ and $\tilde{\mathcal{C}}_{-y}$ contains the remaining. Then, the SupCLR objective is defined by:
\begin{equation}
    \mathcal{L}_{\texttt{SupCLR}}(\mathcal{C}; \mathcal{T})
        := \frac{1}{2B} \sum_{j=1}^{2B}
        \mathcal{L}_{\texttt{con}} (\tilde{x}_j, \tilde{\mathcal{C}}_{y_j} \setminus \{\tilde{x}_j\}, \tilde{\mathcal{C}}_{-y_j}).
    \label{eq:loss-supclr}
\end{equation}
After training the embedding network $f_\theta(x)$ with the SupCLR objective \eqref{eq:loss-supclr}, we train a linear classifier upon $f_\theta(x)$ to model $p_\texttt{SupCLR}(y|x)$.

\textbf{Supervised extension of CSI.} We extend CSI by incorporating the shifting transformations $\mathcal{S}$ into the SupCLR objective: here, we consider a joint label $(y,y^\mathcal{S}) \in \mathcal{Y} \times \mathcal{S}$ of class label $y$ and shifting transformation $y^\mathcal{S}$. 
Then, the \textit{supervised contrasting shifted instances} (sup-CSI) loss is given by:
\begin{equation}
    \mathcal{L}_{\texttt{sup-CSI}}
        := \mathcal{L}_{\texttt{SupCLR}} \left( \bigcup_{S \in \mathcal{S}} \mathcal{C}_S; \mathcal{T} \right),  \quad \text{where} ~~ \mathcal{C}_S:=\{(S(x_i),(y_i,S))\}_{i=1}^B.
    \label{eq:loss-sup-CSI}
\end{equation}
Note that we do not use the auxiliary classification loss $\mathcal{L}_\texttt{cls-SI}$ \eqref{eq:loss-cls-SI}, since the objective already classifies the shifted instances under a \textit{self-label augmented} \citep{lee2020selfsupervised} space $\mathcal{Y} \times \mathcal{S}$.

Upon the learned representation via \eqref{eq:loss-sup-CSI}, we additionally train two linear classifiers: $p_\texttt{CSI}(y|x)$ and $p_\texttt{CSI-joint}(y,y^\mathcal{S}|x)$ that predicts the class labels and joint labels, respectively. We directly apply $s_\texttt{sup}(x)$ for the former $p_\texttt{CSI}(y|x)$. For the latter, on the other hand, we marginalize the joint prediction over the shifting transformation in a similar manner of Section~\ref{sec:method-detect}. Precisely, let $l(x) \in \R^{C \times K}$ be logit values of $p_\texttt{CSI-joint}(y,y^\mathcal{S}|x)$ for $\abs{\mathcal{Y}} = C$ and $\abs{\mathcal{S}} = K$, and $l(x)_k \in \R^C$ be logit values correspond to $p_\texttt{CSI-joint}(y,y^\mathcal{S}=S_k|x)$. Then, the ensembled probability is:
\begin{equation}
    p_\texttt{CSI-ens}(y|x) := \sigma\left( \frac{1}{K} \sum_k l(S_k(x))_k \right),
    \label{eq:score-sup-ens}
\end{equation}
where $\sigma$ denotes the softmax activation. Here, we use $p_\texttt{CSI-ens}$ to compute the confidence $s_\texttt{sup}(x)$. We denote the confidence computed by $p_\texttt{CSI}$ and $p_\texttt{CSI-ens}$ and ``CSI'' and ``CSI-ens'', respectively.

\section{Experiments}
\label{sec:exp}

\begin{table}[t]
\centering
\caption{
AUROC (\%) of various OOD detection methods trained on one-class dataset of (a) CIFAR-10, (b) CIFAR-100 (super-class), and (c) ImageNet-30. For CIFAR-10, we report the means and standard deviations of per-class AUROC averaged over five trials, and the final column indicates the mean AUROC across all the classes. For CIFAR-100 and ImaegeNet-30, we only report the mean AUROC over a single trial. Bold denotes the best results, and $^*$ denotes the values from the reference. See Appendix \ref{appx:oc_more} for additional results, \eg, per-class AUROC on CIFAR-100 and ImageNet-30.
}\label{tab:oc}
\vspace{-0.05in}
\begin{subtable}{\textwidth}
\caption{One-class CIFAR-10} \label{tab:oc-cifar10}
\vspace{-0.05in}
\setlength{\tabcolsep}{2pt} 
\resizebox{\textwidth}{!}{
\begin{tabular}{ll@{\hspace{4pt}}|@{\hspace{4pt}}llllllllll@{\hspace{4pt}}|@{\hspace{3pt}}c}
\toprule
Method & Network & Plane & Car & Bird & Cat & Deer & Dog & Frog & Horse & Ship & Truck & Mean \\
\midrule
OC-SVM$^*$ \citep{scholkopf2000support} & - & 65.6 & 40.9 & 65.3 & 50.1 & 75.2 & 51.2 & 71.8 & 51.2 & 67.9 & 48.5 & 58.8 \\
DeepSVDD$^*$ \citep{ruff2018deep} & LeNet & 61.7 & 65.9 & 50.8 & 59.1 & 60.9 & 65.7 & 67.7 & 67.3 & 75.9 & 73.1 & 64.8 \\
AnoGAN$^*$ \citep{schlegl2017unsupervised} & DCGAN & 67.1 & 54.7 & 52.9 & 54.5 & 65.1 & 60.3 & 58.5 & 62.5 & 75.8 & 66.5 & 61.8 \\
OCGAN$^*$ \citep{perera2019ocgan} & OCGAN & 75.7 & 53.1 & 64.0 & 62.0 & 72.3 & 62.0 & 72.3 & 57.5 & 82.0 & 55.4 & 65.7 \\
Geom$^*$ \citep{golan2018deep} & WRN-16-8 & 74.7 & 95.7 & 78.1 & 72.4 & 87.8 & 87.8 & 83.4 & 95.5 & 93.3 & 91.3 & 86.0 \\
Rot$^*$ \citep{hendrycks2019using_self} & WRN-16-4 & 71.9 & 94.5 & 78.4 & 70.0 & 77.2 & 86.6 & 81.6 & 93.7 & 90.7 & 88.8 & 83.3 \\
Rot+Trans$^*$ \citep{hendrycks2019using_self} & WRN-16-4 & 77.5 & 96.9 & 87.3 & 80.9 & 92.7 & 90.2 & 90.9 & 96.5 & 95.2 & 93.3 & 90.1 \\
GOAD$^*$ \citep{bergman2020classification} & WRN-10-4 & 77.2 & 96.7 & 83.3 & 77.7 & 87.8 & 87.8 & 90.0 & 96.1 & 93.8 & 92.0 & 88.2 \\
Rot \citep{hendrycks2019using_self} & ResNet-18 & 
78.3\stdv{0.2} & 94.3\stdv{0.3} & 86.2\stdv{0.4} & 80.8\stdv{0.6} & 89.4\stdv{0.5} & 89.0\stdv{0.4} & 88.9\stdv{0.4} & 95.1\stdv{0.2} & 92.3\stdv{0.3} & 89.7\stdv{0.3} & 88.4 \\
Rot+Trans \citep{hendrycks2019using_self} & ResNet-18 & 
80.4\stdv{0.3} & 96.4\stdv{0.2} & 85.9\stdv{0.3} & 81.1\stdv{0.5} & 91.3\stdv{0.3} & 89.6\stdv{0.3} & 89.9\stdv{0.3} & 95.9\stdv{0.1} & 95.0\stdv{0.1} & 92.6\stdv{0.2} & 89.8 \\
GOAD \citep{bergman2020classification} & ResNet-18 & 75.5\stdv{0.3} & 94.1\stdv{0.3} & 81.8\stdv{0.5} & 72.0\stdv{0.3} & 83.7\stdv{0.9} & 84.4\stdv{0.3} & 82.9\stdv{0.8} & 93.9\stdv{0.3} & 92.9\stdv{0.3} & 89.5\stdv{0.2} & 85.1 \\
\midrule
CSI (ours) & ResNet-18 & \textbf{89.9}\stdv{0.1} & \textbf{99.1}\stdv{0.0} & \textbf{93.1}\stdv{0.2} & \textbf{86.4}\stdv{0.2} & \textbf{93.9}\stdv{0.1} & \textbf{93.2}\stdv{0.2} & \textbf{95.1}\stdv{0.1} & \textbf{98.7}\stdv{0.0} & \textbf{97.9}\stdv{0.0} & \textbf{95.5}\stdv{0.1} & \textbf{94.3} \\
\bottomrule
\end{tabular}}
\vspace{-0.05in}
\end{subtable}
\begin{subtable}{0.43\textwidth}
\centering\small
\caption{One-class CIFAR-100 (super-class)} \label{tab:oc-cifar100}
\vspace{-0.05in}
\begin{tabular}{llc}
\toprule
Method & Network &  AUROC \\
\midrule
OC-SVM$^*$ \citep{scholkopf2000support} & - & 63.1 \\
Geom$^*$ \citep{golan2018deep} & WRN-16-8 & 78.7 \\
Rot \citep{hendrycks2019using_self} & ResNet-18 & 77.7 \\
Rot+Trans \citep{hendrycks2019using_self} & ResNet-18 & 79.8 \\
GOAD \citep{bergman2020classification} & ResNet-18 & 74.5 \\
CSI (ours) & ResNet-18 & \textbf{89.6} \\
\bottomrule
\end{tabular}
\end{subtable}
\begin{subtable}{0.55\textwidth}
\centering\small
\caption{One-class ImageNet-30} \label{tab:oc-imagenet}
\vspace{-0.05in}
\begin{tabular}{llc}
\toprule
Method & Network &  AUROC \\
\midrule
Rot$^*$ \citep{hendrycks2019using_self} & ResNet-18 & 65.3 \\
Rot+Trans$^*$ \citep{hendrycks2019using_self} & ResNet-18 & 77.9 \\
Rot+Attn$^*$ \citep{hendrycks2019using_self} & ResNet-18 & 81.6 \\
Rot+Trans+Attn$^*$ \citep{hendrycks2019using_self} & ResNet-18 & 84.8 \\
Rot+Trans+Attn+Resize$^*$ \citep{hendrycks2019using_self} & ResNet-18 & 85.7 \\
CSI (ours) & ResNet-18 & \textbf{91.6} \\
\bottomrule
\end{tabular}
\end{subtable}
\end{table}

In Section \ref{sec:exp-main}, we report OOD detection results on unlabeled one-class, unlabeled multi-class, and labeled multi-class datasets. In Section \ref{sec:exp-ablation}, we analyze the effects on various shifting transformations in the context of OOD detection, as well as an ablation study on each component we propose.

\textbf{Setup.}
We use ResNet-18 \citep{he2016deep} architecture for all the experiments. For data augmentations $\mathcal{T}$, we adopt those used by \citet{chen2020simple}: namely, we use the combination of Inception crop \citep{szegedy2015going}, horizontal flip, color jitter, and grayscale. For shifting transformations $\mathcal{S}$, we use the random rotation $0\degree, 90\degree, 180\degree, 270\degree$ unless specified otherwise, as rotation has the highest OOD-ness (see Section~\ref{sec:method-train}) values for natural images, \eg, CIFAR-10 \citep{krizhevsky2009learning}.
However, we remark that the best shifting transformation can be different for other datasets, \eg, Gaussian noise performs better than rotation for texture datasets (see Table~\ref{tab:rot_align} in Section~\ref{sec:exp-ablation}). By default, we train our models from scratch with the training objective in \eqref{eq:loss-csi} and detect OOD samples with the ensembled version of the score in \eqref{eq:score-csi}.

We mainly report the area under the receiver operating characteristic curve (AUROC) as a threshold-free evaluation metric for a detection score. In addition, we report the test accuracy and the expected calibration error (ECE) \citep{naeini2015obtaining, guo2017calibration} for the experiments on labeled multi-class datasets. Here, ECE estimates whether a classifier can indicate when they are likely to be incorrect for test samples (from in-distribution) by measuring the difference between prediction confidence and accuracy. The formal description of the metrics and detailed experimental setups are in Appendix \ref{appx:details}.

\begin{table}[t]
\caption{AUROC (\%) of various OOD detection methods trained on unlabeled (a) CIFAR-10 and (b) ImageNet-30. The reported results are averaged over ﬁve trials, subscripts denote standard deviation, and bold denote the best results. $^*$ denotes the values from the reference.}
\label{tab:unlabel}
\vspace{-0.05in}
\begin{subtable}{\textwidth}
\setlength{\tabcolsep}{3pt} %
\caption{Unlabeled CIFAR-10} \label{tab:unlabel-cifar10}
\vspace{-0.05in}
\resizebox{\textwidth}{!}{
\begin{tabular}{ll@{\hspace{6pt}}lllllll}
\toprule
& & \multicolumn{7}{c}{CIFAR10 $\to$} \\
\cmidrule(l{-1pt}r){3-9}
Method & Network & SVHN & LSUN & ImageNet & LSUN (FIX) & ImageNet (FIX) & CIFAR-100 & Interp. \\
\midrule
Likelihood$^*$ & PixelCNN++ & 8.3 & - & 64.2 & - & - & 52.6 & 52.6 \\
Likelihood$^*$ & Glow & 8.3 & - & 66.3 & - & - & 58.2 & 58.2 \\
Likelihood$^*$ & EBM & 63.0 & - & - & - & - & - & 70.0 \\
Likelihood Ratio$^*$ \citep{ren2019likelihood} & PixelCNN++ & 91.2 & - & - & - & - & - & - \\
Input Complexity$^*$ \citep{serra2020input} & PixelCNN++ & 92.9 & - & 58.9 & - & - & 53.5 & - \\
Input Complexity$^*$ \citep{serra2020input} & Glow & 95.0 & - & 71.6 & - & - & 73.6 & - \\
\midrule
Rot \citep{hendrycks2019using_self} & ResNet-18 & 97.6\stdv{0.2} & 89.2\stdv{0.7} & 90.5\stdv{0.3} & 77.7\stdv{0.3} & 83.2\stdv{0.1} & 79.0\stdv{0.1} & 64.0\stdv{0.3} \\
Rot+Trans \citep{hendrycks2019using_self} & ResNet-18 & 97.8\stdv{0.2} & 92.8\stdv{0.9} & 94.2\stdv{0.7} & 81.6\stdv{0.4} & 86.7\stdv{0.1} & 82.3\stdv{0.2} & 68.1\stdv{0.8} \\
GOAD \citep{bergman2020classification} & ResNet-18 & 96.3\stdv{0.2} & 89.3\stdv{1.5} & 91.8\stdv{1.2} & 78.8\stdv{0.3} & 83.3\stdv{0.1} & 77.2\stdv{0.3} & 59.4\stdv{1.1} \\
CSI (ours) & ResNet-18 & \textbf{99.8}\stdv{0.0} & \textbf{97.5}\stdv{0.3} & \textbf{97.6}\stdv{0.3} & \textbf{90.3}\stdv{0.3} & \textbf{93.3}\stdv{0.1} & \textbf{89.2}\stdv{0.1} & \textbf{79.3}\stdv{0.2} \\
\bottomrule
\end{tabular}}
\vspace{-0.05in}
\end{subtable}
\begin{subtable}{\textwidth}
\caption{Unlabeled ImageNet-30} \label{tab:unlabel-imagenet}
\setlength{\tabcolsep}{5pt} %
\vspace{-0.05in}
\resizebox{\textwidth}{!}{
\begin{tabular}{llcccccccc}
\toprule
 & & \multicolumn{8}{c}{ImageNet-30 $\to$} \\
\cmidrule(lr){3-10}
Method & Network & CUB-200 & Dogs & Pets & Flowers & Food-101 & Places-365 & Caltech-256 & DTD \\
\midrule
Rot \citep{hendrycks2019using_self} & ResNet-18 & 76.5\stdv{0.7} & 77.2\stdv{0.5} & 70.0\stdv{0.5} & 87.2\stdv{0.2} & 72.7\stdv{1.5} & 52.6\stdv{1.4} & 70.9\stdv{0.1} & 89.9\stdv{0.5} \\
Rot+Trans \citep{hendrycks2019using_self} & ResNet-18 & 74.5\stdv{0.5} & 77.8\stdv{1.1} & 70.0\stdv{0.8} & 86.3\stdv{0.3} & 71.6\stdv{1.4} & 53.1\stdv{1.7} & 70.0\stdv{0.2} & 89.4\stdv{0.6} \\
GOAD \citep{bergman2020classification} & ResNet-18 & 71.5\stdv{1.4} & 74.3\stdv{1.6} & 65.5\stdv{1.3} & 82.8\stdv{1.4} & 68.7\stdv{0.7} & 51.0\stdv{1.1} & 67.4\stdv{0.8} & 87.5\stdv{0.8} \\
CSI (ours) & ResNet-18 & \textbf{90.5}\stdv{0.1} & \textbf{97.1}\stdv{0.1} & \textbf{85.2}\stdv{0.2} & \textbf{94.7}\stdv{0.4} & \textbf{89.2}\stdv{0.3} & \textbf{78.3}\stdv{0.3} & \textbf{87.1}\stdv{0.1} & \textbf{96.9}\stdv{0.1} \\
\bottomrule
\end{tabular}}
\end{subtable}
\vspace{-0.1in}
\end{table}

\vspace{-0.05in}
\subsection{Main results}
\vspace{-0.05in}
\label{sec:exp-main}

\textbf{Unlabeled one-class datasets.}
We start by considering the \emph{one-class} setup: here, for a given multi-class dataset of $C$ classes, we conduct $C$ one-class classification tasks, where each task chooses one of the classes as in-distribution while the remaining classes being out-of-distribution. We run our experiments on three datasets, following the prior work \citep{golan2018deep,hendrycks2019using_self,bergman2020classification}: CIFAR-10 \citep{krizhevsky2009learning}, CIFAR-100 labeled into 20 super-classes \citep{krizhevsky2009learning}, and ImageNet-30 \citep{hendrycks2019using_self} datasets. We compare CSI with various prior methods including one-class classifier \citep{scholkopf2000support,ruff2018deep}, reconstruction-based \citep{schlegl2017unsupervised,perera2019ocgan}, and self-supervised \citep{golan2018deep,hendrycks2019using_self,bergman2020classification} approaches. Table \ref{tab:oc} summarizes the results, showing that CSI significantly outperforms the prior methods in all the tested cases. We provide the full, additional results, \eg, class-wise AUROC on CIFAR-100 (super-class) and ImageNet-30, in Appendix \ref{appx:oc_more}.

\textbf{Unlabeled multi-class datasets.}
In this setup, we assume that in-distribution samples are from a specific multi-class dataset without labels, testing on various external datasets as out-of-distribution. We compare CSI on two in-distribution datasets: CIFAR-10 \citep{krizhevsky2009learning} and ImageNet-30 \citep{hendrycks2019using_self}. We consider the following datasets as out-of-distribution: SVHN \citep{netzer2011reading}, resized LSUN and ImageNet \citep{liang2018enhancing}, CIFAR-100 \citep{krizhevsky2009learning}, and linearly-interpolated samples of CIFAR-10 (Interp.) \citep{du2019implicit} for CIFAR-10 experiments, and CUB-200 \citep{welinder2010caltech}, Dogs \citep{khosla2011novel}, Pets \citep{parkhi2012cats}, Flowers \citep{nilsback2006visual}, Food-101 \citep{bossard2014food}, Places-365 \citep{zhou2017places}, Caltech-256 \citep{griffin2007caltech}, and DTD \citep{cimpoi2014describing} for ImageNet-30. We compare CSI with various prior methods, including density-based \citep{du2019implicit,ren2019likelihood,serra2020input} and self-supervised \citep{golan2018deep,bergman2020classification} approaches. 

Table \ref{tab:unlabel} shows the results. Overall, CSI significantly outperforms the prior methods in all benchmarks tested.  We remark that CSI is particularly effective  for detecting hard (\ie, near-distribution) OOD samples, \eg, CIFAR-100 and Interp.~in Table \ref{tab:unlabel-cifar10}. Also, CSI still shows a notable performance in the cases when prior methods often fail, \eg, AUROC of 50\% (\ie, random guess) for Places-365 dataset in Table \ref{tab:unlabel-imagenet}. Finally, we notice that the resized LSUN and ImageNet datasets officially released by \citet{liang2018enhancing} might be misleading to evaluate detection performance for hard OODs: we find that those datasets contain some unintended artifacts, due to incorrect resizing procedure. Such an artifact makes those datasets easily-detectable, \eg, via input statistics. In this respect, we produce and test on their fixed versions, coined LSUN (FIX), and ImageNet (FIX). See Appendix \ref{appx:hard_ood} for details.

\begin{table}[t]
\caption{
Test accuracy (\%), ECE (\%), and AUROC (\%) of confidence-calibrated classifiers trained on labeled (a) CIFAR-10 and (b) ImageNet-30. The reported results are averaged over ﬁve trials for CIFAR-10 and one trial for ImageNet-30. Subscripts denote standard deviation, and bold denote the best results. CSI-ens denotes the ensembled prediction, \ie, 4 times slower (as we use rotation).
}\label{tab:label}
\vspace{-0.05in}
\begin{subtable}{\textwidth}
\caption{Labeled CIFAR-10}
\vspace{-0.05in}
\setlength{\tabcolsep}{4pt} %
\resizebox{\textwidth}{!}{
\begin{tabular}{lccccccccc}
\toprule
& & & \multicolumn{7}{c}{CIFAR10 $\to$} \\
\cmidrule(lr){4-10}
Train method & Test acc. & ECE & SVHN & LSUN & ImageNet & LSUN (FIX) & ImageNet (FIX) & CIFAR100 & Interp. \\
\midrule
Cross Entropy & 93.0\stdv{0.2} & 6.44\stdv{0.2} & 88.6\stdv{0.9} & 90.7\stdv{0.5} & 88.3\stdv{0.6} & 87.5\stdv{0.3} & 87.4\stdv{0.3} & 85.8\stdv{0.3} & 75.4\stdv{0.7} \\
SupCLR \citep{khosla2020supervised} & 93.8\stdv{0.1} & 5.56\stdv{0.1} & 97.3\stdv{0.1} & 92.8\stdv{0.5} & 91.4\stdv{1.2} & 91.6\stdv{1.5} & 90.5\stdv{0.5} & 88.6\stdv{0.2} & 75.7\stdv{0.1} \\
CSI (ours) & 94.8\stdv{0.1} & 4.40\stdv{0.1} & 96.5\stdv{0.2} & 96.3\stdv{0.5} & 96.2\stdv{0.4} & 92.1\stdv{0.5} & 92.4\stdv{0.0} & 90.5\stdv{0.1} & 78.5\stdv{0.2} \\
CSI-ens (ours) & \textbf{96.1}\stdv{0.1} & \textbf{3.50}\stdv{0.1} & \textbf{97.9}\stdv{0.1} & \textbf{97.7}\stdv{0.4} & \textbf{97.6}\stdv{0.3} & \textbf{93.5}\stdv{0.4} & \textbf{94.0}\stdv{0.1} & \textbf{92.2}\stdv{0.1} & \textbf{80.1}\stdv{0.3} \\
\bottomrule
\end{tabular}}
\vspace{-0.05in}
\end{subtable}
\begin{subtable}{\textwidth}
\caption{Labeled ImageNet-30}
\vspace{-0.05in}
\resizebox{\textwidth}{!}{
\begin{tabular}{lcccccccccc}
\toprule
& & & \multicolumn{8}{c}{ImageNet-30 $\to$} \\
\cmidrule(lr){4-11}
Train method & Test acc. & ECE & CUB-200 & Dogs & Pets & Flowers & Food-101 & Places-365 & Caltech-256 & DTD \\
\midrule
Cross Entropy & 94.3 & 5.08 & 88.0 & 96.7 & 95.0 & 89.7 & 79.8 & 90.5 & 90.6 & 90.1 \\
SupCLR \citep{khosla2020supervised} & 96.9 & 3.12 & 86.3 & 95.6 & 94.2 & 92.2 & 81.2 & 89.7 & 90.2 & 92.1 \\
CSI (ours) & 97.0 & 2.61 & 93.4 & 97.7 & 96.9 & 96.0 & 87.0 & 92.5 & 91.9 & 93.7 \\
CSI-ens (ours) & \textbf{97.8} & \textbf{2.19} & \textbf{94.6} & \textbf{98.3} & \textbf{97.4} & \textbf{96.2} & \textbf{88.9} & \textbf{94.0} & \textbf{93.2} & \textbf{97.4} \\
\bottomrule
\end{tabular}}
\end{subtable}
\vspace{-0.1in}
\end{table}

\textbf{Labeled multi-class datasets.}
We also consider the \emph{labeled} version of the above setting: namely, we now assume that every in-distribution sample also contains discriminative label information. We use the same datasets considered in the unlabeled multi-class setup for in- and out-of-distribution datasets. We train our model as proposed in Section~\ref{sec:method-label}, and compare it with those trained by other methods, the cross-entropy and supervised contrastive learning (SupCLR) \citep{khosla2020supervised}. Since our goal is to calibrate the confidence, the maximum softmax probability is used to detect OOD samples (see \citep{hendrycks2017baseline}).

Table \ref{tab:label} shows the results. Overall, CSI consistently improves AUROC and ECE for all benchmarks tested. Interestingly, CSI also improves test accuracy; even our original purpose of CSI is to learn a representation for OOD detection. CSI can further improve the performance by ensembling over the transformations. We also remark that our results on unlabeled datasets (in Table \ref{tab:unlabel}) already show comparable performance to the supervised baselines (in Table \ref{tab:label}).

\vspace{-0.05in}
\subsection{Ablation study}
\vspace{-0.05in}
\label{sec:exp-ablation}

We perform an ablation study on various shifting transformations, training objectives, and detection scores. Throughout this section, we report the mean AUROC values on one-class CIFAR-10.

\begin{figure*}[t]
\centering
\begin{subfigure}{0.13\textwidth}
\includegraphics[width=\textwidth]{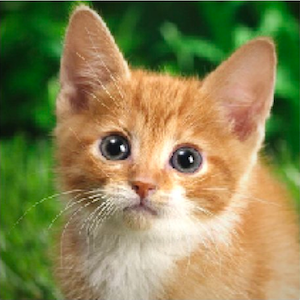}
\caption{Original}\label{fig:ablation_original}
\end{subfigure}
\begin{subfigure}{0.13\textwidth}
\includegraphics[width=\textwidth]{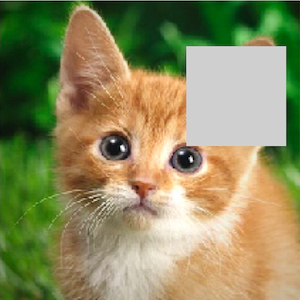}
\caption{Cutout}\label{fig:ablation_cutout}
\end{subfigure}
\begin{subfigure}{0.13\textwidth}
\includegraphics[width=\textwidth]{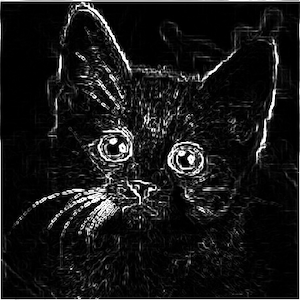}
\caption{Sobel}\label{fig:ablation_sobel}
\end{subfigure}
\begin{subfigure}{0.13\textwidth}
\includegraphics[width=\textwidth]{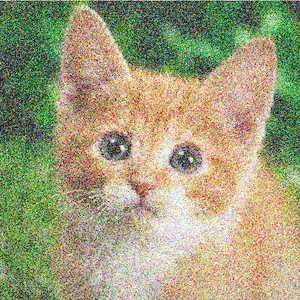}
\caption{Noise}\label{fig:ablation_noise}
\end{subfigure}
\begin{subfigure}{0.13\textwidth}
\includegraphics[width=\textwidth]{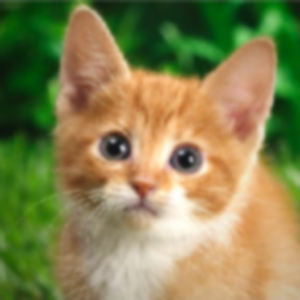}
\caption{Blur}\label{fig:ablation_blur}
\end{subfigure}
\begin{subfigure}{0.13\textwidth}
\includegraphics[width=\textwidth]{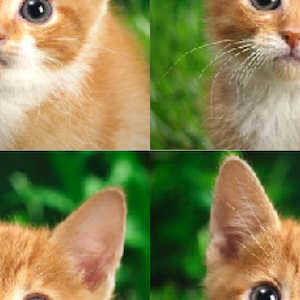}
\caption{Perm}\label{fig:ablation_perm}
\end{subfigure}
\begin{subfigure}{0.13\textwidth}
\includegraphics[width=\textwidth]{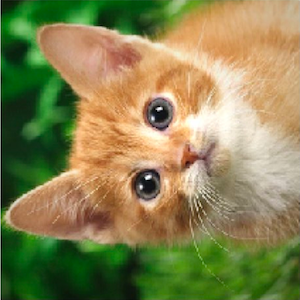}
\caption{Rotate}\label{fig:ablation_rotate}
\end{subfigure}
\caption{
Visualization of the original image and the considered shifting transformations.
}\label{fig:aug_vis}
\centering\small
\captionof{table}{
OOD-ness (\%), \ie, the AUROC between in-distribution vs. transformed samples under the vanilla SimCLR (see Section~\ref{sec:method-train}), of various transformations. The vanilla SimCLR is trained on one-class CIFAR-10 under ResNet-18. Each column denotes the applied transformation.
}\label{tab:ood_aug}
\begin{tabular}{ccccccc}
\toprule
& Cutout & Sobel & Noise & Blur & Perm & Rotate \\
\midrule
OOD-ness & 79.5 & 69.2 & 74.4 & 76.0 & 83.8 & 85.2 \\
\bottomrule
\end{tabular}
\vspace{0.05in}
\centering\footnotesize
\captionof{table}{
Ablation study on various transformations, added or removed from the vanilla SimCLR. ``Align'' and ``Shift'' indicates that the transformation is used as $\mathcal{T}$ and $\mathcal{S}$, respectively. (a) We add a new transformation as an aligned (up) or shifting (down) transformations. (b) We remove (up) or convert-to-shift (down) the transformation from the vanilla SimCLR. All reported values are the mean AUROC (\%) over one-class CIFAR-10, and ``Base'' denotes the vanilla SimCLR.
}\label{tab:ablation-aug}
\vspace{-0.05in}
\addtocounter{table}{-1} %
\begin{subtable}{0.66\textwidth}
\caption{Add transformations}\label{tab:ablation-aug-add}
\begin{tabular}{c|ccccccc}
\toprule
Base && Cutout & Sobel & Noise & Blur & Perm & Rotate \\
\midrule
\multirow{2}{*}{87.9}
& +Align & 84.3 & 85.0 & 85.5 & 88.0 & 73.1 & 76.5 \\
& +Shift & 88.5 & 88.3 & 89.3 & 89.2 & 90.7 & 94.3 \\
\bottomrule
\end{tabular}
\end{subtable}
~\begin{subtable}{0.32\textwidth}
\caption{Remove transformations}\label{tab:ablation-aug-remove}
\begin{tabular}{|cccc}
\toprule
& Crop & Jitter & Gray \\
\midrule
-Align & 55.7 & 78.8 & 78.4 \\
+Shift & - & - & 88.3 \\
\bottomrule
\end{tabular}
\end{subtable}
\vspace{-0.1in}
\end{figure*}

\textbf{Shifting transformation.} 
We measure the OOD-ness (see Section~\ref{sec:method-train}) of transformations, \ie, the AUROC between in-distribution vs. transformed samples under vanilla SimCLR, and the effects of those transformations when used as a shifting transformation. In particular, we consider Cutout \citep{devries2017improved}, Sobel filtering \citep{kanopoulos1988design}, Gaussian noise, Gaussian blur, and rotation \citep{gidaris2018unsupervised}. We remark that these transformations are reported to be ineffective in improving the class discriminative power
of SimCLR \citep{chen2020simple}. We also consider the transformation coined ``Perm'', which randomly permutes each part of the evenly partitioned image.
Intuitively, such transformations commonly \textit{shift} the input distribution, hence forcing them to be \textit{aligned} can be harmful. Figure~\ref{fig:aug_vis} visualizes the considered transformations.

Table \ref{tab:ood_aug} shows AUROC values of the vanilla SimCLR, where the in-distribution samples shifted by the chosen transformation are given as OOD samples. The shifted samples are easily detected: it validates our intuition that the considered transformations \textit{shift} the input distribution. In particular, ``Perm'' and ``Rotate'' are the most distinguishable, which implies they shift the distribution the most. Note that ``Perm'' and ``Rotate'' turns out to be the most effective shifting transformations; it implies that the transformations \textit{shift} the distribution most indeed performs best for CSI.\footnote{We also have tried contrasting \textit{external} OOD samples similarly to \citep{hendrycks2019deep}; however, we find that na\"ively using them in our framework degrade the performance. This is because the contrastive loss also discriminates \textit{within} external OOD samples, which is unnecessary and an additional learning burden for our purpose.}

Besides, we apply the transformation upon the vanilla SimCLR: align the transformed samples to the original samples (\ie, use as $\mathcal{T}$) or consider them as the shifted samples (\ie, use as $\mathcal{S}$). Table \ref{tab:ablation-aug-add} shows that aligning the transformations degrade (or on par) the detection performance, while shifting the transformations gives consistent improvements. We also remove or convert-to-shift the transformation from the vanilla SimCLR in Table \ref{tab:ablation-aug-remove}, and see similar results. We remark that one can further improve the performance by combining multiple shifting transformations (see Appendix \ref{appx:comb_trans}).

\begin{wraptable}[11]{r}{0.42\textwidth}
\vspace{-0.18in}
\centering\small
\caption{OOD-ness (\%) and AUROC (\%) on DTD, where Textile is used for OOD.}%
\vspace{-0.03in}
\label{tab:rot_align}
\begin{minipage}{0.1\textwidth}
\includegraphics[width=\textwidth]{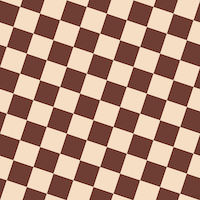}
\end{minipage}
\begin{minipage}{0.1\textwidth}
\includegraphics[width=\textwidth]{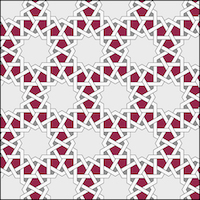}
\end{minipage}
\begin{minipage}{0.1\textwidth}
\includegraphics[width=\textwidth]{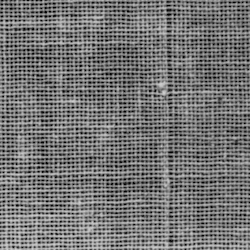}
\end{minipage}
\begin{minipage}{0.1\textwidth}
\includegraphics[width=\textwidth]{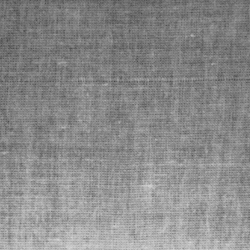}
\end{minipage}
\setlength{\tabcolsep}{5pt}%
\begin{subtable}{0.15\textwidth}
\centering\small
\vspace{0.08in}
\caption{OOD-ness}\label{tab:rot_align_oodness}
\vspace{-0.07in}
\begin{tabular}{cc}
\toprule
Rot. & Noise \\
\midrule
50.6 & \textbf{75.7} \\
\bottomrule
\end{tabular}
\end{subtable}
\begin{subtable}{0.26\textwidth}
\centering\small
\vspace{0.08in}
\caption{AUROC}\label{tab:rot_align_auroc}
\vspace{-0.07in}
\begin{tabular}{ccc}
\toprule
Base & CSI(R) & CSI(N) \\
\midrule
70.3 & 65.9 & \textbf{80.1} \\
\bottomrule
\end{tabular}
\end{subtable}%
\end{wraptable}

\textbf{Data-dependence of shifting transformations.}
We remark that the best shifting transformation depends on the dataset. For example, consider the rotation-invariant datasets: Describable Textures Dataset (DTD) \citep{cimpoi2014describing} and Textile \citep{schulz1996tilda} are in- vs. out-of-distribution, respectively (see Appendix~\ref{appx:texture} for more visual examples). For such datasets, rotation (Rot.) does not shift the distribution, and Gaussian noise (Noise) is more suitable transformation (see Table~\ref{tab:rot_align_oodness}). Table~\ref{tab:rot_align_auroc} shows that CSI using Gaussian noise (``CSI(N)'') indeed improves the vanilla SimCLR (``Base'') while CSI using rotation (``CSI(R)'') degrades instead. This results support our principles on selecting shifting transformations.

\textbf{Linear evaluation.}
We also measure the linear evaluation \citep{kolesnikov2019revisiting}, the  accuracy of a linear classifier to discriminate classes of in-distribution samples. It is widely used for evaluating the quality of (unsupervised) learned representation.
We report the linear evaluation of vanilla SimCLR and CSI (with shifting rotation), trained under unlabeled CIFAR-10. They show comparable results, 90.48\% for SimCLR and 90.19\% for CSI;
CSI is more specialized to learn a representation for OOD detection.

\begin{table}[t]
\centering\footnotesize
\caption{
Ablation study on each component of our proposed (a) training objective and (b) detection score. For (a), we use the corresponding detection score for each training loss; namely, \cref{eq:score-con,eq:score-con-SI,eq:score-cls-SI,eq:score-csi} for \cref{eq:loss-simclr,eq:loss-con-SI,eq:loss-cls-SI,eq:loss-csi}, respectively. For (b), we use the model trained by the final training loss \eqref{eq:loss-csi}. We measure the mean AUROC (\%) values, trained under CIFAR-10 with ResNet-18. Each row indicates the corresponding equation of the given checkmarks, and bold denotes the best results. ``Con.'', ``Cls.'', and ``Ensem.'' denotes contrast, classify, and ensemble, respectively.
}\label{tab:ablation-method}
\vspace{-0.05in}
\begin{subtable}{0.5\textwidth}
\centering
\caption{Training objective}
\label{tab:ablation-train}
\vspace{-0.05in}
\begin{tabular}{rcccc}
\toprule
& SimCLR & Con. & Cls. & AUROC \\
\midrule
$\mathcal{L}_\texttt{SimCLR}$ \eqref{eq:loss-simclr} & \checkmark & - & - & 87.9 \\
$\mathcal{L}_\texttt{con-SI}$ \eqref{eq:loss-con-SI} & \checkmark & \checkmark  & - & 91.6 \\
$\mathcal{L}_\texttt{cls-SI}$ \eqref{eq:loss-cls-SI} & - & - & \checkmark & 88.6 \\
$\mathcal{L}_\texttt{CSI}$ \eqref{eq:loss-csi} & \checkmark & \checkmark & \checkmark & \textbf{94.3} \\
\bottomrule
\end{tabular}
\end{subtable}
\begin{subtable}{0.48\textwidth}
\centering
\caption{Detection score}
\label{tab:ablation-score}
\vspace{-0.05in}
\begin{tabular}{rcccc}
\toprule
& Con. & Cls. & Ensem. & AUROC \\
\midrule
$s_\texttt{con}$ \eqref{eq:score-con} & \checkmark & - & - & 91.3 \\
$s_\texttt{con-SI}$ \eqref{eq:score-con-SI} & \checkmark & - & \checkmark & 93.3 \\
$s_\texttt{cls-SI}$ \eqref{eq:score-cls-SI} & - & \checkmark & \checkmark & 93.8 \\
$s_\texttt{CSI}$ \eqref{eq:score-csi} & \checkmark & \checkmark & \checkmark & \textbf{94.3} \\
\bottomrule
\end{tabular}
\end{subtable}
\vspace{-0.1in}
\end{table}

\textbf{Training objective.}
In Table \ref{tab:ablation-train}, we assess the individual effects of each component that consists of our final training objective \eqref{eq:loss-csi}: namely, we compare the vanilla SimCLR \eqref{eq:loss-simclr}, contrasting shifted instances \eqref{eq:loss-con-SI}, and classifying shifted instances \eqref{eq:loss-cls-SI} losses. For the evaluation of the models of different training objectives \cref{eq:loss-simclr,eq:loss-con-SI,eq:loss-cls-SI,eq:loss-csi}, we use the detection scores defined in \cref{eq:score-con,eq:score-con-SI,eq:score-cls-SI,eq:score-csi}, respectively. We remark that both contrasting and classifying shows better results than the vanilla SimCLR; and combining them (\ie, the final CSI objective \eqref{eq:loss-csi}) gives further improvements, \ie, two losses are complementary.

\textbf{Detection score.}
Finally, Table \ref{tab:ablation-score} shows the effect of each component in our detection score: the vanilla contrastive \eqref{eq:score-con}, contrasting shifted instances \eqref{eq:score-con-SI}, and classifying shifted instances \eqref{eq:score-cls-SI} scores. We ensemble the scores over both $\mathcal{T}$ and $\mathcal{S}$ for \cref{eq:score-con-SI,eq:score-cls-SI,eq:score-csi}, and use a single sample for \cref{eq:score-con}. All the reported values are evaluated from the model trained by the final objective \ref{eq:loss-csi}.
Similar to above, both contrasting and classifying scores show better results than the vanilla contrastive score; and combining them (\ie, the final CSI score \eqref{eq:score-csi}) gives further improvements.

\vspace{-0.06in}
\section{Related work}
\vspace{-0.06in}
\label{sec:related}

\textbf{OOD detection.}
Recent works on unsupervised OOD detection (\ie, no external OOD samples) \citep{hodge2004survey} can be categorized as: (a) density-based \citep{zhai2016deep,nalisnick2019deep,choi2018waic,nalisnick2019detecting,du2019implicit,ren2019likelihood,serra2020input,grathwohl2020your}, (b) reconstruction-based \citep{schlegl2017unsupervised,zong2018deep,deecke2018anomaly,pidhorskyi2018generative,perera2019ocgan,choi2020novelty}, (c) one-class classifier \citep{scholkopf2000support,ruff2018deep}, and (d) self-supervised \citep{golan2018deep,hendrycks2019using_self,bergman2020classification} approaches. Our work falls into (c) the self-supervised approach, as it utilizes the representation learned from self-supervision \citep{gidaris2018unsupervised}. However, unlike prior works \citep{golan2018deep,hendrycks2019using_self,bergman2020classification} focusing on the self-label classification tasks (\eg, predict the angle of the rotated image), we first incorporate \textit{contrastive learning} \citep{chen2020simple} for OOD detection. Concurrently, \citet{winkens2020contrastive} and \citet{liu2020hybrid} report that contrastive learning also improves the OOD detection performance of classifiers \citep{liang2018enhancing,lee2018simple,hendrycks2019using_self}.

\textbf{Confidence-calibrated classifiers.}
Confidence-calibrated classifiers aim to calibrate the prediction confidence (maximum softmax probability), which can be directly used as an uncertainty estimator for both within in-distribution \citep{naeini2015obtaining, guo2017calibration} and in- vs. out-of-distribution \citep{hendrycks2017baseline,lee2018training}. Prior works improved calibration through inference \citep{guo2017calibration} or training \citep{lee2018training} schemes, which are can be jointed applied to our method. Some works design a specific detection score upon the pre-trained classifiers \citep{liang2018enhancing,lee2018simple}, but they only target OOD detection, while ours also consider the in-distribution calibration.

\textbf{Self-supervised learning.}
Self-supervised learning \citep{gidaris2018unsupervised,kolesnikov2019revisiting}, particularly contrastive learning \citep{falcon2020framework} via instance discrimination \citep{wu2018unsupervised}, has shown remarkable success on visual representation learning \citep{he2019momentum,chen2020simple}. However, most prior works focus on the downstream tasks (\eg, classification), and other advantages (\eg, uncertainty or robustness) are rarely investigated \citep{hendrycks2019using_self,kim2020adversarial}. Our work, concurrent with \citep{liu2020hybrid,winkens2020contrastive}, first verifies that contrastive learning is also effective for OOD detection. In particular, we find that the shifting transformations, which were known to be harmful and unused for the standard contrastive learning \citep{chen2020simple}, can help OOD detection. This observation provides new considerations for selecting transformations, \ie, which transformation should be used for positive or negative \citep{tian2020makes,xiao2020should}.

We further provide a more comprehensive survey and discussions with prior works in Appendix~\ref{appx:related}.
\vspace{-0.06in}
\section{Conclusion}
\vspace{-0.06in}
\label{sec:conclusion}

We propose a simple yet effective method named contrasting shifted instances (CSI), which extends the power of contrastive learning for out-of-distribution (OOD) detection problems. CSI demonstrates outstanding performance under various OOD detection scenarios. We believe our work would guide various future directions in OOD detection and self-supervised learning as an important baseline.

\section*{Acknowledgements}
This work was supported by Institute of Information \& Communications Technology Planning \& Evaluation (IITP) grant funded by the Korea government (MSIT) (No.2019-0-00075, Artificial Intelligence Graduate School Program (KAIST) and No.2017-0-01779, A machine learning and statistical inference framework for explainable artificial intelligence). We thank Sihyun Yu, Chaewon Kim, Hyuntak Cha, Hyunwoo Kang, and Seunghyun Lee for helpful feedback and suggestions.

\section*{Broader Impact}

This paper is focused on the subject of \textit{out-of-distribution (OOD)} (or novelty, anomaly) detection, which is an essential ingredient for building safe and reliable intelligent systems \citep{amodei2016concrete}. We expect our results to have two consequences for academia and broader society.

\textbf{Rethinking representation for OOD detection.}
In this paper, we demonstrate that the representation for classification (or other related tasks, measured by linear evaluation \citep{kolesnikov2019revisiting}) can be different from the representation for OOD detection. In particular, we verify that the ``hard'' augmentations, thought to be harmful for contrastive representation learning \citep{chen2020simple}, can be helpful for OOD detection. Our observation raises new questions for both representation learning and OOD detection: (a) representation learning researches should also report the OOD detection results as an evaluation metric, (b) OOD detection researches should more investigate the specialized representation.

\textbf{Towards reliable intelligent system.}
The intelligent system should be robust to the potential dangers of uncertain environments (\eg, financial crisis \citep{taylor2009black}) or malicious adversaries (\eg, cybersecurity \citep{kruegel2003anomaly}). Detecting outliers is also related to human safety (\eg, medical diagnosis \citep{caruana2015intelligible} or autonomous driving \citep{eykholt2018robust}), and has a broad range of industrial applications (\eg, manufacturing inspection \citep{lucke2008smart}). However, the system can be stuck into \textit{confirmation bias}, \ie, ignore new information with a myopic perspective. We hope the system to balance the exploration and exploitation of the knowledge.

\nocite{lee2017confident,yun2020regularizing}
\bibliographystyle{abbrvnat}
\bibliography{ref}

\appendix

\onecolumn
\clearpage
\begin{center}{\bf {\LARGE Appendix}}
\end{center}
\begin{center}{\bf {\Large CSI: Novelty Detection via Contrastive Learning\\on Distributionally Shifted Instances}}
\end{center}

\vspace{0.1in}
\section{Experimental details}
\label{appx:details}

\textbf{Training details.}
We use ResNet-18 \citep{he2016deep} as the base encoder network $f_\theta$ and 2-layer multi-layer perceptron with 128 embedding dimension as the projection head $g_{\phi}$. All models are trained by minimizing the final loss $\mathcal{L}_{\texttt{CSI}}$ \eqref{eq:loss-csi} with a temperature of $\tau=0.5$. We follow the same optimization step of SimCLR \citep{chen2020simple}. For optimization, we train CSI with 1,000 epoch under LARS optimizer \citep{you2017large} with weight decay of $1\mathrm{e}{-6}$ and momentum with 0.9. For the learning rate scheduling, we use linear warmup \citep{goyal2017accurate} for early 10 epochs until learning rate of 1.0 and decay with cosine decay schedule without a restart \citep{loshchilov2016sgdr}. We use batch size of 512 for both vanilla SimCLR and ours: where the batch is given by $\mathcal{B}$ for vanilla SimCLR and the aggregated one $\bigcup_{S \in \mathcal{S}} \mathcal{B}_S$ for ours. Furthermore, we use global batch normalization (BN) \citep{ioffe2015batch}, which shares the BN parameters (mean and variance) over the GPUs in distributed training.

For supervised contrastive learning (SupCLR) \citep{khosla2020supervised} and supervised CSI, we select the best temperature from $\{0.07, 0.5\}$: SupCLR recommend 0.07 but 0.5 was better in our experiments. For training the encoder $f_\theta$, we use the same optimization scheme as above, except using 700 for the epoch. For training the linear classifier, we train the model for 100 epochs with batch size 128, using stochastic gradient descent with momentum 0.9. The learning rate starts at 0.1 and is dropped by a factor of 10 at 60\%, 75\%, and 90\% of the training progress.

\textbf{Data augmentation details.} We use SimCLR augmentations: Inception crop \citep{szegedy2015going}, horizontal flip, color jitter, and grayscale for random augmentations $\mathcal{T}$, and rotation as shifting transformation $\mathcal{S}$. The detailed description of the augmentations are as follows:

\begin{itemize}
\item \textbf{Inception crop.} Randomly crops the area of the original image with uniform distribution 0.08 to 1.0. After the crop, cropped image are resized to the original image size.
\item \textbf{Horizontal flip.} Flips the image horizontally with 50\% of probability. 
\item \textbf{Color jitter.} Change the hue, brightness, and saturation of the image. We transform the RGB (red, green, blue) image into an HSV (hue, saturation, value) image format and add noise to the HSV channels. We apply color jitter with 80\% of probability.
\item \textbf{Grayscale.} Convert into a gray image. Randomly apply a grayscale with 20\% of probability.
\item \textbf{Rotation.} We use rotation as $\mathcal{S}$, the shifting transformation, $\{0\degree, 90\degree, 180\degree, 270\degree\}$. For a given batch $\mathcal{B}$, we apply each rotation degree to obtain the new batch for CSI: $\bigcup_{S \in \mathcal{S}} \mathcal{B}_S$.
\end{itemize}

\begin{figure*}[h]
\centering
\begin{subfigure}{0.19\textwidth}
\includegraphics[width=\textwidth]{figures/original.png}
\caption{Original}
\end{subfigure}
~\begin{subfigure}{0.19\textwidth}
\includegraphics[width=\textwidth]{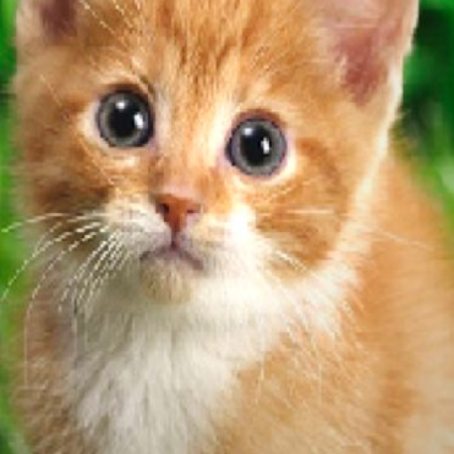}
\caption{Inception crop}
\end{subfigure}
\centering
~\begin{subfigure}{0.19\textwidth}
\includegraphics[width=\textwidth]{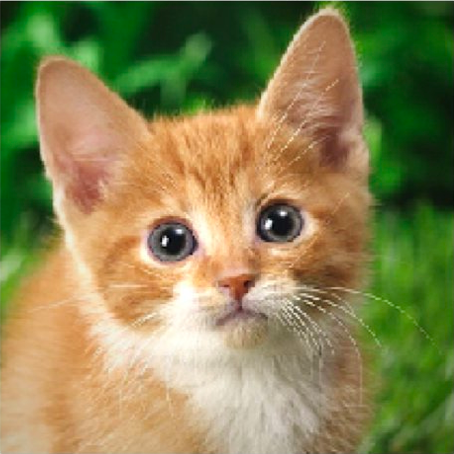}
\caption{Horizontal flip}
\end{subfigure}
\centering
~\begin{subfigure}{0.19\textwidth}
\includegraphics[width=\textwidth]{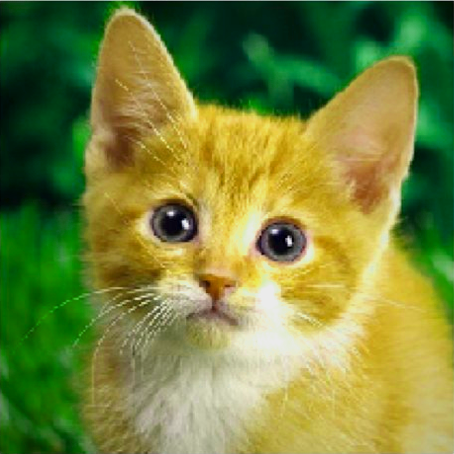}
\caption{Color jitter}
\end{subfigure}
\centering
~\begin{subfigure}{0.19\textwidth}
\includegraphics[width=\textwidth]{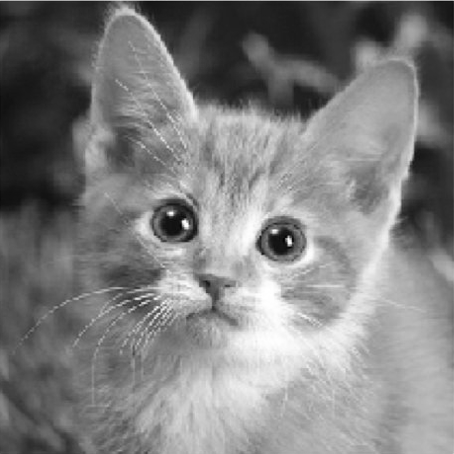}
\caption{Grayscale}
\end{subfigure}
\caption{
Visualization of original image and SimCLR augmentations.
}\label{fig:simclr_aug}
\end{figure*}

\clearpage
\textbf{Dataset details.} For one-class datasets, we train one class of CIFAR-10 \citep{krizhevsky2009learning}, CIFAR-100 (super-class) \citep{krizhevsky2009learning}, and ImageNet-30 \citep{hendrycks2019using_self}. CIFAR-10 and  CIFAR-100 consist of 50,000 training and 10,000 test images with 10 and 20 (super-class) image classes, respectively. ImageNet-30 contains 39,000 training and 3,000 test images with 30 image classes.

For unlabeled and labeled multi-class datasets, we train ResNet with CIFAR-10 and ImageNet-30. For CIFAR-10, out-of-distribution (OOD) samples are as follows: SVHN \citep{netzer2011reading} consists of 26,032 test images with 10 digits, resized LSUN \citep{liang2018enhancing} consists of 10,000 test images of 10 different scenes, resized ImageNet \citep{liang2018enhancing} consists of 10,000 test images with 200 images classes from a subset of full ImageNet dataset, Interp. consists of 10,000 test images of linear interpolation of CIFAR-10 test images, and LSUN (FIX), ImageNet (FIX) consists of 10,000 test images, respectively with following details in Appendix \ref{appx:hard_ood}. For multi-class ImageNet-30, OOD samples are as follows: CUB-200 \citep{welinder2010caltech}, Stanford Dogs \citep{khosla2011novel}, Oxford Pets \citep{parkhi2012cats}, Oxford Flowers \citep{nilsback2006visual}, Food-101 \citep{bossard2014food} without the ``hotdog'' class to avoid overlap, Places-365 \citep{zhou2017places} with small images (256 * 256) validation set, Caltech-256 \citep{griffin2007caltech}, and Describable Textures Dataset (DTD) \citep{cimpoi2014describing}. Here, we randomly sample 3,000 images to balance with the in-distribution test set.

\textbf{Evaluation metrics.} 
For evaluation, we measure the two metrics that each measures (a) the effectiveness of the proposed score in distinguishing in- and out-of-distribution images, (b) the confidence calibration of softmax classifier. 

\begin{itemize}
\item \textbf{Area under the receiver operating characteristic curve (AUROC).} Let TP, TN, FP, and FN denote true positive, true negative, false positive and false negative, respectively. The ROC curve is a graph plotting true positive rate = TP / (TP+FN) against the false positive rate = FP / (FP+TN) by varying a threshold.
\item \textbf{Expected calibration error (ECE).} For a given test data $\{(x_n,y_n)\}_{n=1}^N$, we group the predictions into $M$ interval bins (each of size $1/M$). Let $B_m$ be the set of indices of samples whose prediction confidence falls into the interval $(\frac{m-1}{M},\frac{m}{M}]$. Then, the expected calibration error (ECE) \citep{naeini2015obtaining, guo2017calibration} is follows:
\begin{equation}
    \mathrm{ECE} = \sum_{m=1}^M \frac{|B_m|}{N}|\mathrm{acc}(B_m) - \mathrm{conf}(B_m)|,
    \label{eq:ece}
\end{equation}
where $\mathrm{acc}(B_m)$ is accuracy of $B_m$: $\mathrm{acc}(B_m)=\frac{1}{|B_m|}\sum_{i\in B_m} \mathbbm{1}_{\{y_i=\arg\max_y p(y|x_i)\}}$ where $\mathbbm{1}$ is indicator function and $\mathrm{conf}(B_m)$ is confidence of $B_m$: $\mathrm{conf}(B_m)=\frac{1}{|B_m|}\sum_{i\in B_m}q(x_i)$ where $q(x_i)$ is the confidence of data $x_i$. 
\end{itemize}

\clearpage
\section{Detailed review on related work}
\label{appx:related}

\subsection{OOD detection}

Out-of-distribution (OOD) detection is a classic and essential problem in machine learning, studied under different names, \eg, novelty or anomaly detection \citep{hodge2004survey}. In this paper, we primarily focus on \textit{unsupervised} OOD detection, which is arguably the most traditional and popular setup in the field \citep{scholkopf2000support}. In this setting, the detector can only access in-distribution samples while required to identify unseen OOD samples. There are other settings, \eg, semi-supervised setting - the detector can access a small subset of out-of-distribution samples \citep{hendrycks2019deep,ruff2020deep}, or supervised setting - the detector knows the target out-of-distribution, but we do not consider those settings in this paper. We remark that the unsupervised setting is the most practical and challenging scenario since there are \textit{infinitely} many cases for out-of-distribution, and it is often not possible to have such external data.

Most recent works can be categorized as: (a) density-based \citep{zhai2016deep,nalisnick2019deep,choi2018waic,nalisnick2019detecting,du2019implicit,ren2019likelihood,serra2020input,grathwohl2020your}, (b) reconstruction-based \citep{schlegl2017unsupervised,zong2018deep,deecke2018anomaly,pidhorskyi2018generative,perera2019ocgan,choi2020novelty}, (c) one-class classifier \citep{scholkopf2000support,ruff2018deep,ruff2020deep}, and (d) self-supervised \citep{golan2018deep,hendrycks2019using_self,bergman2020classification} approaches. We note that there are more extensive literature on this topic, but we mainly focus on the recent work based on deep learning. Brief description for each method are as follows:
\begin{itemize}
    \item \textbf{Density-based methods.} Density-based methods are one of the most classic and principled approaches for OOD detection. Intuitively, they directly use the likelihood of the sample as the detection score. However, recent studies reveal that the likelihood is often not the best metric - especially for deep neural networks with complex datasets \citep{nalisnick2019deep}. Several work thus proposed modified scores, \eg, typicality \citep{nalisnick2019detecting}, WAIC \citep{choi2018waic}, likelihood ratio \citep{ren2019likelihood}, input complexity \citep{serra2020input}, or unnormalized likelihood (\ie, energy) \citep{du2019implicit,grathwohl2020your}.
    \item \textbf{Reconstruction-based methods.} Reconstruction-based approach is another popular line of research for OOD detection. It trains an encoder-decoder network that reconstructs the training data in an unsupervised manner. Since the network would less generalize for unseen OOD samples, they use the reconstruction loss as a detection score. Some works utilize auto-encoders \citep{zong2018deep,pidhorskyi2018generative} or generative adversarial networks \citep{schlegl2017unsupervised,deecke2018anomaly,perera2019ocgan}.
    \item \textbf{One-class classifiers.} One-class classifiers are also a classic and principled approach for OOD detection. They learn a decision boundary of in- vs.\ out-of-distribution samples by giving some margin covering the in-distribution samples \citep{scholkopf2000support}. Recent works have shown that the one-class classifier is effective upon the deep representation \citep{ruff2018deep}.
    \item \textbf{Self-supervised methods.} Self-supervised approaches are a relatively new technique based on the rich representation learned from self-supervision \citep{gidaris2018unsupervised}. They train a network with a pre-defined task (\eg, predict the angle of the rotated image) on the training set, and use the generalization error to detect OOD samples. Recent self-supervised approaches show outstanding results on various OOD detection benchmark datasets \citep{golan2018deep,hendrycks2019using_self,bergman2020classification}.
\end{itemize}

Our work falls into (c) the self-supervised approach \citep{golan2018deep,hendrycks2019using_self,bergman2020classification}. However, unlike prior work focusing on the self-label classification tasks (\eg, rotation \citep{gidaris2018unsupervised}) which trains an auxiliary classifier to predict the transformation applied to the sample, we first incorporate \textit{contrastive learning} \citep{chen2020simple} for OOD detection. To that end, we design a novel detection score utilizing the unique characteristic of contrastive learning, \eg, the features in the projection layer learned by cosine similarity. We also propose a novel self-supervised training scheme that further improves the representation for OOD detection. Nevertheless, we acknowledge that the prior work largely inspired our work. For instance, the classifying shifted instances loss \eqref{eq:loss-cls-SI} follows the form of auxiliary classifiers \citep{hendrycks2019using_self}, which gives further improvement upon our novel contrasting shifted instances loss \eqref{eq:loss-con-SI}.

Concurrently, \citet{winkens2020contrastive} and \citet{liu2020hybrid} report the similar observations that contrastive learning also improves the OOD detection performance of classifiers \citep{liang2018enhancing,lee2018simple,hendrycks2019using_self}. \citet{winkens2020contrastive} jointly train a classifier with the SimCLR \citep{chen2020simple} objective and use the Mahalanobis distance \citep{lee2018simple} as a detection score. \citet{liu2020hybrid} approximates JEM \citep{grathwohl2020your} (a joint model of classifier and energy-based model \citep{du2019implicit}) by a combination of classification and contrastive loss and use density-based detection scores \citep{grathwohl2020your}. In contrast to both work, we mainly focus on the \textit{unlabeled OOD} setting (although we also discuss the confident-calibrated classifiers). Here, we design a novel detection score, since how to utilize the contrastive representation (which is learned in an unsupervised manner) for OOD detection have not been explored before.

\subsection{Confidence-calibrated classifiers}

Another line of research is on confidence-calibrated classifiers \citep{hendrycks2017baseline}, which relaxes the overconfidence issue of the classifiers. There are two types of calibration: (a) \textit{in-distribution} calibration \citep{naeini2015obtaining, guo2017calibration}, that aligns the uncertainty and the actual accuracy, measured by ECE, and (b) \textit{out-of-distribution} detection \citep{hendrycks2017baseline,lee2018training}, that reduces the uncertainty of OOD samples, measured by AUROC. Note that the goal of confidence-calibrated classifiers is to regularize the prediction. Hence, the softmax probability is used for all three tasks: classification, in-distribution calibration, and out-of-distribution detection. Namely, the detection score is given by the prediction confidence (or maximum softmax probability) \citep{hendrycks2017baseline}. Prior works improved calibration through inference (temperature scaling) \citep{guo2017calibration} or training (regularize predictions of OOD samples) \citep{lee2018training} schemes, which can be jointly applied to our method. Some works design a specific detection score upon the pre-trained classifiers \citep{liang2018enhancing,lee2018simple}, but they only target OOD detection, while ours also consider the in-distribution calibration.

\subsection{Self-supervised learning}

Self-supervised learning \citep{gidaris2018unsupervised,kolesnikov2019revisiting} has shown remarkable success in learning representations. In particular, contrastive learning \citep{falcon2020framework} via instance discrimination \citep{wu2018unsupervised} show the state-of-the-art results on visual representation learning \citep{he2019momentum,chen2020simple}. However, most prior works focus on improving the downstream task performance (\eg, classification), and other advantages of self-supervised learning (\eg, uncertainty or robustness) are rarely investigated \citep{hendrycks2019using_self,kim2020adversarial}. Our work, concurrent with \citep{liu2020hybrid,winkens2020contrastive}, first verifies that contrastive learning is also effective for OOD detection.

Furthermore, we find that the shifting transformations, which were known to be harmful and unused for the standard contrastive learning \citep{chen2020simple}, can help OOD detection. This observation provides new considerations for selecting transformations, \ie, which transformation should be used for positive or negative \citep{tian2020makes,xiao2020should}. Specifically, \citet{tian2020makes} claims the optimal views (or transformations) of the \textit{positive} pairs should minimize the mutual information while keeping the task-relevant information. It suggests that the shifting transformation may not contain the information for classification, but may contain OOD detection information when used for the \textit{negative} pairs. \citet{xiao2020should} suggests a framework that automatically learns whether the transformation should be positive or negative. One could consider incorporating our principle on shifting transformation (\ie, OOD-ness); OOD detection could be another evaluation metric for the learned representations.

\clearpage
\section{Additional one-class OOD detection results}
\label{appx:oc_more}

Table \ref{tab:oc-confusion} presents the confusion matrix of AUROC values of our method on one-class CIFAR-10 datasets, where bold denotes the hard pairs. The results align with the human intuition that `car' is confused to `ship' and `truck', and `cat' is confused to `dog'.

Table \ref{tab:oc-cifar100-full} presents the OOD detection results of various methods on one-class CIFAR-100 (super-class) datasets, for all 20 super-classes. Our method outperforms the prior methods for all classes.

Table \ref{tab:oc-imagenet-full} presents the OOD detection results of our method on one-class ImageNet-30 dataset, for all 30 classes. Our method consistently performs well for all classes.

\begin{table}[h]
\vspace{-0.05in}
\centering\small
\caption{
Confusion matrix of AUROC (\%) values of our method on one-class CIFAR-10. The row and column indicates the in-distribution and OOD class, respectively, and the final column indicates the mean value. Bold denotes the values under 80\%, which implies the hard pair.
}\label{tab:oc-confusion}
\begin{tabular}{lcccccccccc|c}
\toprule
& Plane & Car & Bird & Cat & Deer & Dog & Frog & Horse & Ship & Truck & Mean \\
\midrule
Plane & - & \textbf{74.1} & 95.8 & 98.4 & 94.9 & 98.0 & 96.2 & 90.1 & \textbf{79.6} & 82.8 & 90.0 \\
Car   & 99.3 & - & 99.9 & 99.9 & 99.8 & 99.9 & 99.8 & 99.7 & 98.7 & 95.0 & 99.1 \\
Bird  & 91.1 & 97.5 & - & 97.3 & 87.0 & 92.5 & 96.1 & 83.2 & 96.4 & 98.0 & 93.2 \\
Cat   & 91.9 & 91.5 & 90.3 & - & 83.3 & \textbf{67.0} & 89.6 & \textbf{79.0} & 92.8 & 91.9 & 86.4 \\
Deer  & 95.7 & 98.4 & 94.9 & 96.6 & - & 94.7 & 98.7 & \textbf{69.0} & 97.4 & 98.8 & 93.8 \\
Dog   & 97.9 & 98.5 & 95.5 & 90.3 & 88.1 & - & 96.8 & \textbf{76.6} & 98.6 & 98.3 & 93.4 \\
Frog  & 93.6 & 92.3 & 94.6 & 96.1 & 96.8 & 96.3 & - & 95.2 & 94.4 & 97.3 & 95.2 \\
Horse & 99.3 & 99.5 & 99.0 & 99.3 & 94.2 & 97.4 & 99.8 & - & 99.7 & 99.4 & 98.6 \\
Ship  & 96.6 & 91.2 & 99.5 & 99.7 & 99.4 & 99.7 & 99.5 & 99.3 & - & 96.6 & 97.9 \\
Truck & 96.2 & \textbf{72.3} & 99.4 & 99.5 & 99.1 & 99.4 & 98.7 & 98.3 & 96.2 & - & 95.5 \\
\bottomrule
\end{tabular}
\end{table}
\begin{table*}[h]
\vspace{-0.05in}
\centering\small
\caption{
AUROC (\%) values of various OOD detection methods trained on one-class CIFAR-100 (super-class). Each row indicates the results of the selected super-class, and the final row indicates the mean value. $^*$ denotes the values from the reference, and bold denotes the best results.
}\label{tab:oc-cifar100-full}
\vspace{-0.1in}
\resizebox{\textwidth}{!}{
\begin{tabular}{ccccccccccc}
\toprule
& OC-SVM$^*$ & DAGMM$^*$ & DSEBM$^*$ & ADGAN$^*$ & Geom$^*$ & Rot & Rot+Trans & GOAD & CSI (ours) \\
\midrule
0 & 68.4 & 43.4 & 64.0 & 63.1 & 74.7 & 78.6 & 79.6 & 73.9 & \textbf{86.3} \\
1 & 63.6 & 49.5 & 47.9 & 64.9 & 68.5 & 73.4 & 73.3 & 69.2 & \textbf{84.8} \\
2 & 52.0 & 66.1 & 53.7 & 41.3 & 74.0 & 70.1 & 71.3 & 67.6 & \textbf{88.9} \\
3 & 64.7 & 52.6 & 48.4 & 50.0 & 81.0 & 68.6 & 73.9 & 71.8 & \textbf{85.7} \\
4 & 58.2 & 56.9 & 59.7 & 40.6 & 78.4 & 78.7 & 79.7 & 72.7 & \textbf{93.7} \\
5 & 54.9 & 52.4 & 46.6 & 42.8 & 59.1 & 69.7 & 72.6 & 67.0 & \textbf{81.9} \\
6 & 57.2 & 55.0 & 51.7 & 51.1 & 81.8 & 78.8 & 85.1 & 80.0 & \textbf{91.8} \\
7 & 62.9 & 52.8 & 54.8 & 55.4 & 65.0 & 62.5 & 66.8 & 59.1 & \textbf{83.9} \\
8 & 65.6 & 53.2 & 66.7 & 59.2 & 85.5 & 84.2 & 86.0 & 79.5 & \textbf{91.6} \\
9 & 74.1 & 42.5 & 71.2 & 62.7 & 90.6 & 86.3 & 87.3 & 83.7 & \textbf{95.0} \\
10 & 84.1 & 52.7 & 78.3 & 79.8 & 87.6 & 87.1 & 88.6 & 84.0 & \textbf{94.0} \\
11 & 58.0 & 46.4 & 62.7 & 53.7 & 83.9 & 76.2 & 77.1 & 68.7 & \textbf{90.1} \\
12 & 68.5 & 42.7 & 66.8 & 58.9 & 83.2 & 83.3 & 84.6 & 75.1 & \textbf{90.3} \\
13 & 64.6 & 45.4 & 52.6 & 57.4 & 58.0 & 60.7 & 62.1 & 56.6 & \textbf{81.5} \\
14 & 51.2 & 57.2 & 44.0 & 39.4 & 92.1 & 87.1 & 88.0 & 83.8 & \textbf{94.4} \\
15 & 62.8 & 48.8 & 56.8 & 55.6 & 68.3 & 69.0 & 71.9 & 66.9 & \textbf{85.6} \\
16 & 66.6 & 54.4 & 63.1 & 63.3 & 73.5 & 71.7 & 75.6 & 67.5 & \textbf{83.0} \\
17 & 73.7 & 36.4 & 73.0 & 66.7 & 93.8 & 92.2 & 93.5 & 91.6 & \textbf{97.5} \\
18 & 52.8 & 52.4 & 57.7 & 44.3 & 90.7 & 90.4 & 91.5 & 88.0 & \textbf{95.9} \\
19 & 58.4 & 50.3 & 55.5 & 53.0 & 85.0 & 86.5 & 88.1 & 82.6 & \textbf{95.2} \\
\midrule
Mean & 63.1 & 50.6 & 58.8 & 55.2 & 78.7 & 77.7 & 79.8 & 74.5 & \textbf{89.6} \\
\bottomrule
\end{tabular}}
\end{table*}

\begin{table*}[h]
\vspace{-0.05in}
\centering\small
\caption{
AUROC (\%) values of our method trained on one-class ImageNet-30. The first and third row indicates the selected class, and the second and firth row indicates the corresponding results.
}\label{tab:oc-imagenet-full}
\vspace{-0.1in}
\resizebox{\textwidth}{!}{
\begin{tabular}{ccccccccccccccc}
\toprule
0 & 1 & 2 & 3 & 4 & 5 & 6 & 7 & 8 & 9 & 10 & 11 & 12 & 13 & 14 \\
\midrule
85.9 & 99.0 & 99.8 & 90.5 & 95.8 & 99.2 & 96.6 & 83.5 & 92.2 & 84.3 & 99.0 & 94.5 & 97.1 & 87.7 & 96.4 \\
\midrule
\midrule
15 & 16 & 17 & 18 & 19 & 20 & 21 & 22 & 23 & 24 & 25 & 26 & 27 & 28 & 29 \\
\midrule
84.7 & 99.7 & 75.6 & 95.2 & 73.8 & 94.7 & 95.2 & 99.2 & 98.5 & 82.5 & 89.7 & 82.1 & 97.2 & 82.1 & 97.6 \\
\bottomrule
\end{tabular}}
\end{table*}

\clearpage
\section{Ablation study on random augmentation}
\label{appx:test_aug}

We verify that ensembling the scores over the random augmentations $\mathcal{T}$ improves OOD detection. However, na\"ive random sampling from the entire $\mathcal{T}$ is often sample inefficient. We find that choosing a proper subset $\mathcal{T}_\texttt{control} \subset \mathcal{T}$ improves the performance for given number of samples. Specifically, we choose $\mathcal{T}_\texttt{control}$ as the set of the \textit{most common} samples. For example, the size of the cropping area is sampled from $\mathcal{U}[0.08,1]$ for uniform distribution $\mathcal{U}$ during training. Since the rare samples, \eg, area near $0.08$ increases the noise, we only use the samples with size $(0.08 + 1) / 2 = 0.54$ during inference. Table \ref{tab:test_aug} shows random sampling from the controlled set often gives improvements.

\begin{table}[h]
\centering\small
\caption{
AUROC (\%) values of our method for different number of random augmentations, under one-class (OC-) CIFAR-10 and CIFAR-100 (super-class). The values are averaged over classes. Random augmentations over the controlled set show the best performance.
}\label{tab:test_aug}
\begin{tabular}{cccc}
\toprule
\# of samples & Controlled & OC-CIFAR-10 & OC-CIFAR-100 \\
\midrule
4 & - & 92.22 & 87.36 \\
40 & - & 94.13 & 89.51 \\
40 & \checkmark & \textbf{94.31} & \textbf{89.55} \\
\bottomrule
\end{tabular}
\end{table}

\section{Efficient computation of \eqref{eq:score-con} via coreset}
\label{appx:coreset}

One can reduce the computation and memory cost of the contrastive score \eqref{eq:score-con} by selecting a proper subset, \ie, \textit{coreset}, of the training samples. To this end, we run K-means clustering \citep{macqueen1967some} on the normalized features $W_m := z(x_m) / \norm{z(x_m)}$ using cosine similarity as a metric. Then, we use the center of each cluster as the coreset. For contrasting shifted instances \eqref{eq:loss-cls-SI}, we choose the coreset for each shifting transformation $S$. Table \ref{tab:coreset} shows the results for various coreset sizes, given by a ratio from the full training samples. Keeping only a few (\eg, 1\%) samples is sufficient.

\begin{table}[h]
\centering\small
\caption{
AUROC (\%) values of our method for various corset sizes (\% of training samples), under one-class (OC-) CIFAR-10, CIFAR-100 (super-class), and ImageNet-30. The values are averaged over classes. Keeping only a few (\eg, 1\%) samples shows sufficiently good results.
}\label{tab:coreset}
\begin{tabular}{rccc}
\toprule
Coreset (\%) & OC-CIFAR-10 & OC-CIFAR-100 & OC-ImageNet-30 \\
\midrule
1\%   & 94.22 & 89.27 & 91.06 \\
10\%  & 94.30 & 89.46 & 91.51 \\
100\% & 94.31 & 89.55 & 91.63 \\
\bottomrule
\end{tabular}
\end{table}

\clearpage
\section{Ablation study on the balancing terms}
\label{appx:balance}

We study the effects of the balancing terms $\lambda_S^\texttt{con}$, $\lambda_S^\texttt{cls}$ in Section \ref{sec:method-detect}. To this end, we compare of our final loss \eqref{eq:loss-csi}, without (w/o) and with (w/) the balancing terms $\lambda_S^\texttt{con}$ and $\lambda_S^\texttt{cls}$. When not using the balancing terms, we set $\lambda_S^\texttt{con} = \lambda_S^\texttt{cls} = 1$ for all $S$. We follow the experimental setup of Table \ref{tab:oc}, \eg, use rotation for the shifting transformation. We run our experiments on CIFAR-10, CIFAR-100 (super-class), and ImageNet-30 datasets. Table \ref{tab:balance} shows that the balancing terms gives a consistent improvement. CIFAR-10 do not show much gain since all $\lambda_S^\texttt{con}$ and $\lambda_S^\texttt{cls}$ show similar values; in contrast, CIFAR-100 (super-class) and ImageNet-30 show large gain since they varies much.

\begin{table}[h]
\centering\small
\caption{
AUROC (\%) values of our method without (w/o) and with (w/) balancing terms, under one-class (OC-) CIFAR-10, CIFAR-100 (super-class), and ImageNet-30. The values are averaged over classes, and bold denotes the best results. Balancing terms give consistent improvements.
}\label{tab:balance}
\begin{tabular}{lccc}
\toprule
& OC-CIFAR-10 & OC-CIFAR-100 & OC-ImageNet-30 \\
\midrule
CSI (w/o balancing) & 94.28 & 89.00 & 91.04 \\
CSI (w/ balancing) & \textbf{94.31} & \textbf{89.55} & \textbf{91.63} \\
\bottomrule
\end{tabular}
\end{table}

\section{Combining multiple shifting transformations}
\label{appx:comb_trans}

We find that combining multiple shifting transformations: given two transformations $\mathcal{S}_1$ and $\mathcal{S}_2$, use $\mathcal{S}_1 \times \mathcal{S}_2$ as the combined shifting transformation, can give further improvements. Table \ref{tab:comb_trans} shows that combining ``Noise'', ``Blur'', and ``Perm'' to ``Rotate'' gives additional gain. We remark that one can investigate the better combination; we choose rotation for our experiments due to its simplicity.

\begin{table}[h]
\centering\small
\caption{
AUROC (\%) values of our method under various shifting transformations. Combining ``Noise'', ``Blur'', and ``Perm'' to ``Rotate'' gives additional gain.
}\label{tab:comb_trans}
\begin{tabular}{rc|cccccccc}
\toprule
& Base & Noise & Blur & Perm & Rotate & Rotate+Noise & Rotate+Blur & Rotate+Perm  \\
\midrule
AUROC & 87.89 & 89.29 & 89.15 & 90.68 & 94.31 & \textbf{94.65} & \textbf{94.66} & \textbf{94.60} \\
\bottomrule
\end{tabular}
\end{table}

\clearpage
\section{Discussion on the features of the contrastive score \eqref{eq:score-con}}
\label{appx:norm}

We find that the two features: a) the \textit{cosine similarity} to the nearest training sample in $\{x_m\}$, \ie, $\max_m \mathrm{sim}(z(x_m), z(x))$, and (b) the \emph{feature norm} of the representation, \ie, $\norm{z(x)}$, are important features for detecting OOD samples under the SimCLR representation.

In this section, we first demonstrate the properties of the two features under vanilla SimCLR. While we use the vanilla SimCLR to validate they are general properties of SimCLR, we remark that our training scheme (see Section \ref{sec:method-train}) further improves the discrimination power of the features. Next, we verify that cosine similarity and feature norm are \textit{complementary}, that combining both features (\ie, $s_\texttt{con}$ \eqref{eq:score-con}) give additional gain. For the latter one, we use our final training loss to match the reported values in prior experiments, but we note that the trend is consistent among the models.

\begin{figure*}[h]
\centering
\begin{subfigure}{0.32\textwidth}
\includegraphics[width=\textwidth]{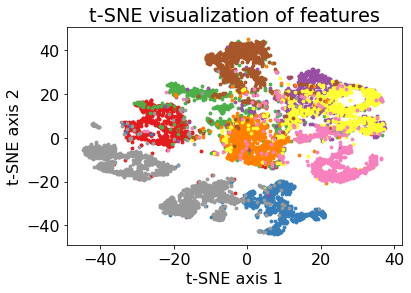}
\caption{t-SNE visualization}\label{fig:sim-1}
\end{subfigure}
~\begin{subfigure}{0.32\textwidth}
\includegraphics[width=\textwidth]{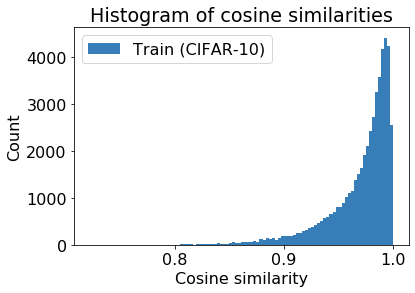}
\caption{Similarities (train)}\label{fig:sim-2}
\end{subfigure}
~\begin{subfigure}{0.32\textwidth}
\includegraphics[width=\textwidth]{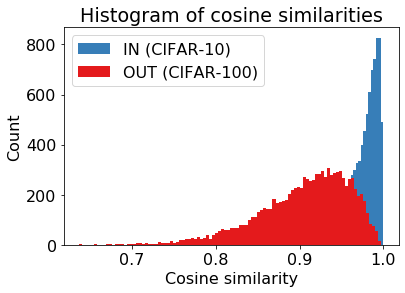}
\caption{Similarities (test)}\label{fig:sim-3}
\end{subfigure}
\caption{
Plots for cosine similarity.
}\label{fig:sim}
\end{figure*}

First, we demonstrate the effect of cosine similarity for OOD detection. To this end, we train vanilla SimCLR using CIFAR-10 and CIFAR-100 and in- and out-of-distribution datasets. Since SimCLR attracts the same image with different augmentations, it learns to cluster similar images; hence, it shows good discrimination performance measured by linear evaluation \citep{chen2020simple}. Figure \ref{fig:sim-1} presents the t-SNE \citep{maaten2008visualizing} plot of the normalized features that each color denote different class. Even though SimCLR is trained in an unsupervised manner, the samples of the same classes are gathered.

Figure \ref{fig:sim-2} and Figure \ref{fig:sim-3} presents the histogram of the cosine similarities from the nearest training sample (\ie, $\max_m \mathrm{sim}(z(x_m), z(x))$), for training and test datasets, respectively. For the training set, we choose the second nearest sample since the nearest one is itself. One can see that training samples are concentrated, even though contrastive learning pushes the different samples. It complements the results of Figure \ref{fig:sim-1}. For test sets, the in-distribution samples show a similar trend with the training samples. However, the OOD samples are farther from the training samples, which implies that the cosine similarity is an effective feature to detect OOD samples.

\begin{figure*}[h]
\centering
\begin{subfigure}{0.32\textwidth}
\includegraphics[width=\textwidth]{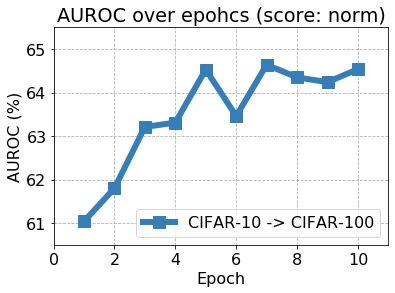}
\caption{Trend of AUROC}\label{fig:norm-1}
\end{subfigure}
~\begin{subfigure}{0.32\textwidth}
\includegraphics[width=\textwidth]{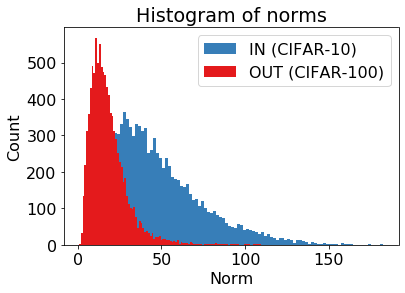}
\caption{Histogram of norms}\label{fig:norm-2}
\end{subfigure}
~\begin{subfigure}{0.32\textwidth}
\includegraphics[width=\textwidth]{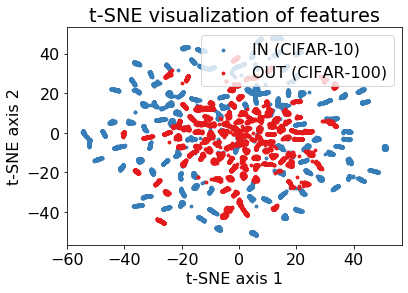}
\caption{t-SNE visualization}\label{fig:norm-3}
\end{subfigure}
\caption{
Plots for feature norm.
}\label{fig:norm}
\end{figure*}

\clearpage
Second, we demonstrate that the feature norm is a discriminative feature for OOD detection. Following the prior setting, we use CIFAR-10 and CIFAR-100 for in- and out-of-distribution datasets, respectively. Figure \ref{fig:norm-1} shows that the discriminative power of feature norm improves as the training epoch increases. We observe that this phenomenon consistently happens over models and settings; the contrastive loss makes the norm of in-distribution samples relatively larger than OOD samples. Figure \ref{fig:norm-2} shows the norm of CIFAR-10 is indeed larger than CIFAR-100, under the final model.

This is somewhat unintuitive since the SimCLR uses the \textit{normalized} features to compute the loss \eqref{eq:loss-con}. To understand this phenomenon, we visualize the t-SNE \citep{maaten2008visualizing} plot of the feature space in Figure \ref{fig:norm-3}, randomly choosing 100 images from both datasets. We randomly augment each image for 100 times for better visualization. One can see that in-distribution samples tend to be spread out over the large sphere, while OOD samples are gathered near center.\footnote{t-SNE plot \textit{does not} tell the true behavior of the original feature space, but it may give some intuition.} Also, note that the same image with different augmentations are highly clustered, while in-distribution samples are slightly more assembled.\footnote{We also try the local variance of the norm as a detection score. It also works well, but the norm is better.}

We suspect that increasing the norm may be an \textit{easier} way to maximize cosine similarity between two vectors: instead of directly reducing the feature distance of two augmented samples, one can also increase the overall norm of the features to reduce the \textit{relative} distance of two samples.

Finally, we verify that cosine similarity (sim-only) and feature norm (norm-only) are complementary: combining them (sim+norm) gives additional improvements. Here, we use the model trained by our final objective \eqref{eq:loss-csi}, and follow the inference scheme of the main experiments (see Table \ref{tab:ablation-method}). Table \ref{tab:sim_norm} shows AUROC values under sim-only, norm-only, and sim+norm scores. Using only sim or norm already shows good results, but combining them shows the best results.

\begin{table}[h]
\centering\small
\caption{
AUROC (\%) values for sim-only, norm-only, and sim+norm (\ie, contrastive \eqref{eq:score-con}) scores, under one-class (OC-) CIFAR-10, CIFAR-100 (super-class), and ImageNet-30. The values are averaged over classes. Using both sim and norm features shows the best results.
}\label{tab:sim_norm}
\begin{tabular}{rccc}
\toprule
& OC-CIFAR-10 & OC-CIFAR-100 & OC-ImageNet-30 \\
\midrule
Sim-only & 90.12 & 86.57 & 83.18 \\
Norm-only & 92.70 & 87.71 & 88.56 \\
Sim+Norm & \textbf{93.32} & \textbf{88.79} & \textbf{89.32} \\
\bottomrule
\end{tabular}
\end{table}

\clearpage
\section{Rethinking OOD detection benchmarks}
\label{appx:hard_ood}

\begin{figure*}[h]
\centering
\includegraphics[width=0.24\textwidth]{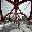}
~\includegraphics[width=0.24\textwidth]{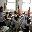}
~\includegraphics[width=0.24\textwidth]{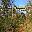}
~\includegraphics[width=0.24\textwidth]{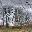}
\vspace{-0.05in}
\caption{
Current benchmark datasets: resized LSUN (left two) and ImageNet (right two).
}\label{fig:easy_ood}
\vspace{0.1in}
\centering
\includegraphics[width=0.24\textwidth]{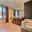}
~\includegraphics[width=0.24\textwidth]{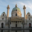}
~\includegraphics[width=0.24\textwidth]{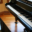}
~\includegraphics[width=0.24\textwidth]{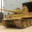}
\vspace{-0.05in}
\caption{
Proposed datasets: LSUN (FIX) (left two) and ImageNet (FIX) (right two).
}\label{fig:hard_ood}
\end{figure*}

We find that resized LSUN and ImageNet \citep{liang2018enhancing}, one of the most popular benchmark datasets for OOD detection, are visually far from in-distribution datasets (commonly, CIFAR \citep{krizhevsky2009learning}). Figure~\ref{fig:easy_ood} shows that resized LSUN and ImageNet contain artificial noises, produced by broken image operations.\footnote{It is also reported in \url{https://twitter.com/jaakkolehtinen/status/1258102168176951299}.} It is problematic since one can detect such datasets with simple data statistics, without understanding semantics from neural networks. To progress OOD detection research one step further, one needs more \textit{hard} or \textit{semantic} OOD samples that cannot be easily detected by data statistics.

To verify this, we propose a simple detection score that measures the \textit{input smoothness} of an image. Intuitively, noisy images would have a higher variation in input space than natural images. Formally, let $x^{(i,j)}$ be the $i$-th value of the vectorized image $x \in \R^{H W K}$. Here, we define the \textit{neighborhood} $\mathcal{N}$ as the set of spatially connected pairs of pixel indices. Then, the \textit{total variation} distance is given by
\begin{equation}
    \mathrm{TV}(x) = \sum_{i,j \in \mathcal{N}} \norm{x^{(i)} - x^{(j)}}_2^2.
    \label{eq:smoothness}
\end{equation}
Then, we define the \textit{smoothness score} as the difference of total variation from the training samples:
\begin{equation}
    s_\texttt{smooth}(x) := \abs{\mathrm{TV}(x) - \frac{1}{M} \sum_m \mathrm{TV}(x_m)}.
    \label{eq:score_soothness}
\end{equation}
Table \ref{tab:easy-ood} shows that this simple score detects current benchmark datasets surprisingly well.

To address this issue, we construct new benchmark datasets, using a fixed resize operation\footnote{We use PyTorch \code{torchvision.transforms.Resize()} operation.}, hence coined LSUN (FIX) and ImageNet (FIX). For LSUN (FIX), we randomly sample 1,000 images from every ten classes of the training set of LSUN. For ImageNet (FIX), we randomly sample 10,000 images from the entire training set of ImageNet-30, excluding ``airliner'', ``ambulance'', ``parking-meter'', and ``schooner'' classes to avoid overlapping with CIFAR-10.\footnote{We provide the datasets and data generation code in \url{https://github.com/alinlab/CSI}.} Figure \ref{fig:hard_ood} shows that the new datasets are more visually realistic than the former ones (Figure \ref{fig:easy_ood}). Also, Table \ref{tab:easy-ood} shows that the fixed datasets are not detected by the simple data statistics \eqref{eq:score_soothness}. We believe our newly produced datasets would be a stronger benchmark for hard or semantic OOD detection for future researches.

\begin{table}[ht]
\centering\small
\caption{AUROC (\%) values using the smoothness score \eqref{eq:score_soothness}, under unlabeled CIFAR-10. Bold denotes the values over 80\%, which implies the dataset is easily detected.
}\label{tab:easy-ood}
\begin{tabular}{ccccccc}
\toprule
\multicolumn{7}{c}{CIFAR10 $\to$} \\
\midrule
SVHN & LSUN & ImageNet & LSUN (FIX) & ImageNet (FIX) & CIFAR-100 & Interp. \\
\midrule
\textbf{85.88} & \textbf{95.70} & \textbf{90.53} & 44.13 & 52.76 & 52.14 & 66.17 \\
\bottomrule
\end{tabular}
\end{table}

\section{Additional examples of rotation-invariant images}
\label{appx:texture}

\begin{figure*}[h]
\centering
\includegraphics[width=0.24\textwidth]{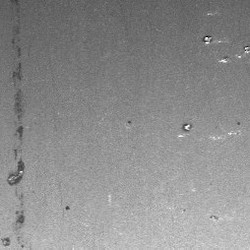}
~\includegraphics[width=0.24\textwidth]{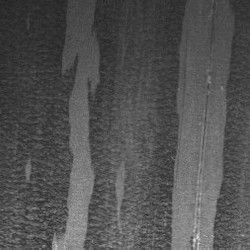}
~\includegraphics[width=0.24\textwidth]{figures/textile_1.png}
~\includegraphics[width=0.24\textwidth]{figures/textile_2.png}
\caption{
Examples of steel (left two) and textile (right two) images.
}\label{fig:texture}
\vspace{0.2in}
\includegraphics[width=0.24\textwidth]{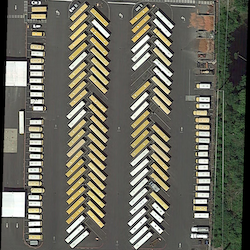}
~\includegraphics[width=0.24\textwidth]{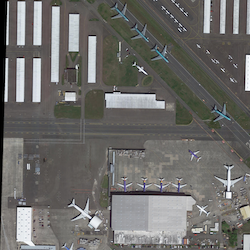}
~\includegraphics[width=0.24\textwidth]{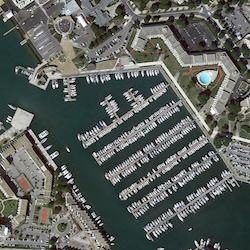}
~\includegraphics[width=0.24\textwidth]{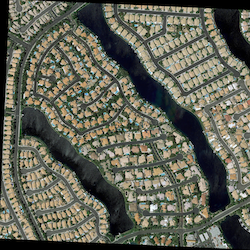}
\caption{
Examples of aerial images.
}\label{fig:aerial}
\end{figure*}

We provide additional examples of rotation-invariant images (see Table~\ref{tab:rot_align} in Section \ref{sec:exp-ablation}). Those image commonly appear in real-world scenarios since many practical applications deal with non-natural images, \eg, manufacturing - steel \citep{severstal} or textile \citep{schulz1996tilda} for instance, or aerial \citep{xia2018dota} images. Figure~\ref{fig:texture} and Figure~\ref{fig:aerial} visualizes the samples of manufacturing and aerial images, respectively. 

\end{document}